\documentclass[10pt,twocolumn,letterpaper]{article}

\usepackage{wacv}      %

\usepackage{times}
\usepackage{epsfig}
\usepackage{graphicx}
\usepackage{amsmath}
\usepackage{amssymb}
\usepackage{booktabs}
\usepackage{enumitem}
\usepackage{xcolor}
\usepackage{multirow}
\usepackage{multicol}
\usepackage{overpic}
\usepackage{caption}
\usepackage{amsmath}
\usepackage[accsupp]{axessibility}

\usepackage[pagebackref,breaklinks,colorlinks]{hyperref}

\usepackage[capitalize]{cleveref}
\crefname{section}{Sec.}{Secs.}
\Crefname{section}{Section}{Sections}
\Crefname{table}{Table}{Tables}
\crefname{table}{Tab.}{Tabs.}

\def\confName{WACV}
\def\confYear{2024}

\begin{document}

\title{FIRe \includegraphics[width=12pt]{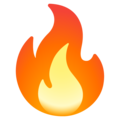}: Fast Inverse Rendering using Directional and Signed Distance Functions
}
\author{
Tarun Yenamandra\textsuperscript{1}
\and
Ayush Tewari\textsuperscript{2}
\and
Nan Yang\textsuperscript{1}
\and
Florian Bernard\textsuperscript{3}
\and
Christian Theobalt\textsuperscript{4}
\and
Daniel Cremers\textsuperscript{1}
\and 
{ \small {}\textsuperscript{1}TU Munich, MCML  
 {}\textsuperscript{2}Massachusetts Institute of Technology {}\textsuperscript{3}University of Bonn {}\textsuperscript{4}MPI Informatics, SIC}
}
\twocolumn[{
\renewcommand\twocolumn[1][]{#1}
\maketitle
\begin{center}
    \centering
    \vspace{-1cm}
    \captionsetup{type=figure}
    \begin{overpic}[width=0.95\linewidth,,tics=5,trim=4 4 4 4,clip]
        {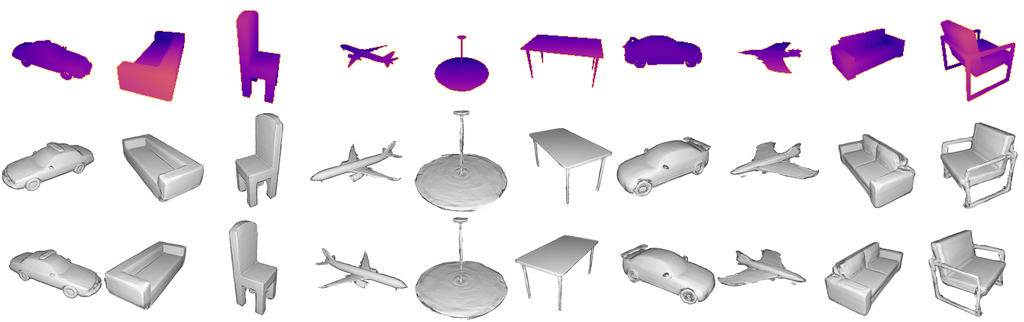}
        \put(-3.5,23){\rotatebox{90}{\small Input}}
        \put(-1.5,23){\rotatebox{90}{\small Depth}}
        \put(-3.5,1){\rotatebox{90}{\small Reconstructions (DDF)}}
        \put(-1.5,12){\rotatebox{90}{\small View 1}}
        \put(-1,2){\rotatebox{90}{\small View 2}}
    \end{overpic}
    \captionof{figure}{We propose a novel neural scene representation  based on directional distance function (DDF), which enables us to replace sphere tracing for rendering images from a signed distance function (SDF) model. We learn the SDF and DDF models on a class of 3D shapes. During inference, given a depth map (top row), we reconstruct 3D shapes by means of our proposed algorithm (FIRe) which is $15$ times faster (per iteration) and more accurate than competing methods. In the last two rows, we show images of reconstructions rendered using our DDF model with just a \textbf{single network evaluation per ray}.
    }
    \label{fig:teaser}
\end{center}%
}]
\maketitle

\begin{abstract}
\vspace{-0.5cm}
Neural 3D implicit representations learn priors that are useful for diverse applications, such as single- or multiple-view 3D reconstruction. A major downside of existing approaches while rendering an image is that they require evaluating the network multiple times per camera ray so that the high computational time forms a bottleneck for downstream applications. We address this problem by introducing a novel neural scene representation that we call the directional distance function (DDF). To this end, we learn a  signed distance function (SDF) along with our DDF model to represent a class of shapes. Specifically, our DDF is defined on the unit sphere and predicts the distance to the surface along any given direction. Therefore, our DDF allows rendering images with just a single network evaluation per camera ray. Based on our DDF, we present a novel fast algorithm (FIRe) to reconstruct 3D shapes given a posed depth map. We evaluate our proposed method on 3D reconstruction from single-view depth images, where we empirically show that our algorithm reconstructs 3D shapes more accurately and it is more than $15$ times faster (per iteration) than competing methods. 

\end{abstract}

\section{Introduction}
The field of generating 3D shapes~\cite{park2019deepsdf,mescheder2019onet, yenamandra2021i3dmm, peng2020convolutional} has seen unprecedented growth in the recent past due to novel neural network architectures. Yet, there are many open challenges for generating realistic 3D shapes, such as data availability and 3D shape representations. Further, using the 3D generative models for accurately reconstructing 3D shapes given partial observations such as depth maps or point clouds is still in the early stages.

Implicit scene representations have proven to be the most suitable data representations for generating 3D surfaces using deep neural networks. Among others, signed distance functions (SDFs) are commonly used. SDFs represent a 3D shape as the level-set of a function, ${\{x\in \mathbb{R}^3 \,| \, f(x)=0\}}$. At every point in space, the SDF of a 3D shape evaluates to the minimum distance to the surface. The sign indicates if the point is inside or outside the shape. For rendering an image of the shape represented by SDFs, one must perform a line search along each camera ray to find the distance to the surface. Sphere tracing~\cite{hart1996spheretracing} accelerates this process for SDFs by exploiting the minimum distance property of SDFs. Inverse rendering is the process of optimizing for the shape and other properties from one or many images~\cite{yariv2020idr}. 

Substantial progress has been made towards single shape or scene reconstruction~\cite{mildenhall2020nerf,yariv2020idr} from dense multi-view images using inverse rendering. Some of the models~\cite{takikawa2021nglod,chen2022tensorf,chan2022eg3d,mueller2022instant,chabra2020deep} trade off memory for speed enabling real-time rendering. However, these models cannot be used as priors as they reconstruct a single scene. Further, the major focus of these methods is to generate novel views of a scene rather than reconstruct geometry. It is an open problem to train such models to represent different shapes with accurate geometry.

In contrast, a 3D generative model learns a conditional implicit function of shapes. In addition to a 3D point, a 3D generative model accepts a latent code as an input to represent different shapes. DeepSDF~\cite{park2019deepsdf} learns a class of shapes using an autodecoding framework. Models trained on many shapes can be used as priors to reconstruct shapes from partial observations, such as images, at test time. We use inverse rendering to optimize for the latent code of a generative model during inference. However, for each optimization step, we need to render an image by sphere tracing through a neural implicit representation as done in DIST~\cite{liu2020dist}.

 \textbf{Novel Scene Representation:}
In this paper, we propose to accelerate inverse rendering algorithms with learned models by avoiding sphere tracing at each iteration of the algorithms. Towards that, we propose a novel scene representation called directional distance function (DDF). We propose to use DDF along with the signed distance function (SDF). We assume that the 3D shapes that our models represent are inside the unit sphere. While the SDF is defined everywhere, our DDF is defined on the surface of the unit sphere. Our DDF model learns to predict the distance to the object's surface along rays cast in all directions from the unit sphere's surface. DDF has two output components - the directional distance, and the probability of the ray hitting the surface. The learned DDF model accelerates inverse rendering algorithms by reducing the number of network evaluations required to find the object surface to $1$ for each iteration.
 
 \textbf{Enabling Fast Inverse Rendering:} We propose a shape optimization algorithm that utilizes our proposed neural representation to reconstruct the 3D shape given partial observations, such as single view depth. As our DDF model replaces the sphere tracing algorithm, our algorithm is $15.5\times$ faster than competing methods.
Our contributions are as follows.
\begin{enumerate}[leftmargin=*]
    \item A novel neural scene representation, DDF defined on the unit sphere, for rendering images from our SDF model during inference with $1$ forward pass through the model.
    \item An algorithm to reconstruct 3D shapes from single view depth maps using our DDF and SDF models, which is $15.5\times$ per iteration faster than competing methods.
\end{enumerate}

\section{Related Work}
In the following, we introduce relevant papers from different domains. \\

\textbf{Implicit Representations:} Implicit shape representations, in particular SDFs, have been studied for decades as they can represent shapes with arbitrary topology
~\cite{Faugeras-Keriven-98,leventon2002statistical,cremers2006dynamical,kohlberger20064d,sturm2013copyme3d,ricci1973constructive,muraki1991volumetric,sommer2022gradientsdf,kobbelt2001featureSensitive}. 
Recently, neural network-based implicits~\cite{park2019deepsdf, mescheder2019onet} have proven to be a compact way to represent SDFs. DeepSDF~\cite{park2019deepsdf} learns the SDF values using an autodecoder architecture conditioned on a learned latent code set to generate different 3D shapes. OccupancyNet~\cite{mescheder2019onet} and IMNet~\cite{chen2018implicitdecoder} learn object surfaces as decision boundaries using an autoencoder architecture. Instead of encoding global priors, local priors~\cite{jiang2020local,chibane2020ifnet,peng2020convolutional,xu2019disn} have been explored to handle large-scale scenes and more detailed representations. However, these rely on generating large feature grids using neural networks. Further, for each new downstream task such as reconstructing from a depth map, or even handling a change in input image resolution, they need to train a new encoder. Differentiable rendering-based methods~\cite{niemeyer2020dvr,sitzmann2019scene} alleviate the need for 3D supervision by learning from images. Pixel features~\cite{yu2020pixelnerf,saito2019pifu,saito2020pifuhd} have been used to condition implicit representations for novel view synthesis given a single image. However, they cannot model the geometry of the objects satisfactorily. Other novel view synthesis~\cite{sitzmann2021lfns} methods suffer from similar problems. Applications of implicit representation on human~\cite{tiwari22posendf,saito2020pifuhd,saito2019pifu,tiwari21neuralgif}, face~\cite{yenamandra2021i3dmm}, and hair~\cite{saito20183d,yenamandra2021i3dmm} modeling are also explored and have achieved superior results to classical methods.\\

\textbf{Directional Distance Prediction:}
Recent efforts towards predicting the occupancy density distribution along the rays~\cite{piala2021terminerf}, or, alternatively, a region along the ray instead of distance~\cite{neff2021donerf}, have proven to accelerate volumetric rendering. However, they only model single objects and they still need to perform local sampling for volumetric rendering. We note CPDDF~\cite{aumentado2021pddf}, PRIF~\cite{feng2022prif}, NeuralODF~\cite{houchens2022neuralodf}, and SDDF~\cite{zobeidi2021sddf} as our contemporary works, which propose to use DDF as a standalone representation. However, unlike these methods, we use both SDF and DDF for high-quality geometric details while defining the DDF only on the surface of the unit sphere.

\textbf{Single View 3D Reconstruction:} 
Single-view reconstruction is generally an ill-posed problem, general solutions~\cite{gkioxari2019meshrcnn,toeppe_et_al_cvpr13,Wu_2020_CVPR,ye2021joint} exploit low-level geometric or photometric properties, whereas shape-specific methods~\cite{tewari17MoFA,bogo2016keepitsmpl,wang2020directshape,niemeyer2020dvr,xu2019disn, yu2020pixelnerf, saito2019pifu,sitzmann2021lfns} solve the problem using learned priors~\cite{blanz1999bfm,loper2015smpl,park2019deepsdf}.
The closest to our algorithm is DIST~\cite{liu2020dist}, which reconstructs 3D shapes given a depth map. However, it requires multiple evaluations per ray, whereas ours needs only a single evaluation.
\section{Method} \label{sec:method}
We learn the two neural representations, DDF and SDF. The DDF model represents the distance to the surface of an object from a point on the unit sphere along a given direction, and the probability of the ray hitting the surface. This helps avoid the computationally intensive sphere tracing step for rendering images, especially for solving image-based inverse problems such as 3D reconstruction.
In the following, we first introduce our representation, followed by our network architecture, and then our proposed algorithm.

\subsection{Directional Distance Representation}
Our 3D shape representation consists of two components: (i) a distance $d$ to the surface of an object along a given direction $r$ from a point $p$ on the surface of the unit sphere, called the directional distance (DDF) and  (ii) the signed distance $s$ at every point inside the unit sphere enclosing the object (SDF).

The directional distance $d$ and signed distance $s$ are related as follows - outside the surface of the object the signed distance is positive and the value is the minimum distance between a given point $x \in \mathbb{R}^3$ and the object surface (in any given direction), i.e.,
\begin{align*}
    \text{SDF}(x) = \min_r \text{DDF}(x,r),
\end{align*}
where $s = \text{SDF}(x)$ is the signed distance at the point ${x \in \mathbb{R}^3}$, $r\in \mathbb{S}^2$ is a given direction from $x$ pointing towards the surface of the object.

In our proposed directional distance representation, we learn to predict the DDF on the unit sphere $p \in \mathbb{S}^2$ along with a ray-hitting probability $\sigma \in [0,1]$. Further, the value of SDF at the distance predicted along a hitting direction $r$ must be $0$, or
\begin{align}
    \text{SDF}(p+d_{\sigma=1}r) &= 0, d_{\sigma=1} = \text{DDF}(p,r). \label{eq:SDFAtDDF}
\end{align}
In the following, we introduce our neural model that learns to represent this function along with SDF, and show how we exploit the relationship defined in Eq.~\eqref{eq:SDFAtDDF}.

\subsection{Network}
\begin{figure}[ht]
\centering
\includegraphics[width=\linewidth]{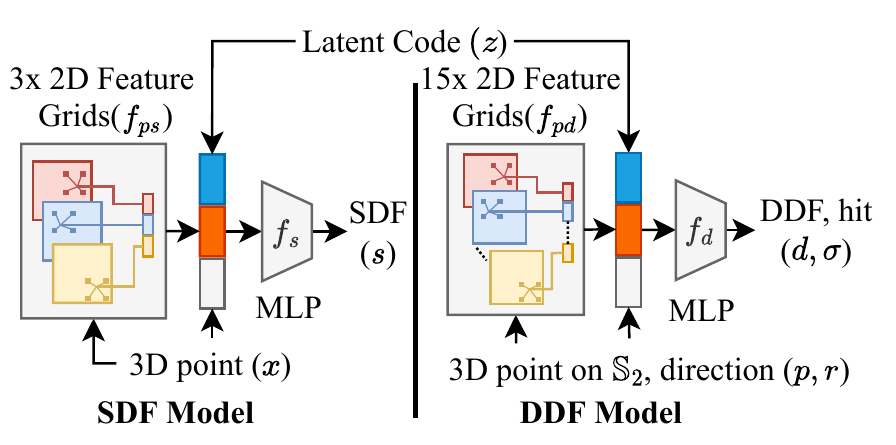}
\vspace{-7mm}
\caption{SDF and DDF Models: Our SDF model generalizes with high-dimensional feature inputs from $3$ 2D feature grids ($f_{ps}$) by sampling from the grid with bilinear interpolation given a 3D point ($x \in \mathbb{R}^3$) and a latent code ($z$). Similarly, our DDF model takes as input a point on the unit sphere and a direction ($(p,r)\in \mathbb{R}^6$) along with a latent code ($z$) to generalize with features from $15$ 2D feature grids for each shape category. The SDF and DDF models have a shared latent space for each shape category.}
\vspace{-7mm}
\label{fig:ddfpipeline}
\end{figure} 

Our model consists of two shape representations, SDF and DDF. For both, we use neural networks conditioned on latent codes to represent multiple shapes. Further, we make our generative model of 3D shapes viable to represent shapes with much higher accuracy. We achieve this by conditioning them on high-dimensional features that are sampled from learned feature planes for each shape category, as shown in Fig.~\ref{fig:ddfpipeline}. Our network architecture is inspired by Pi-GAN's~\cite{chanmonteiro2020pigan} implementation\footnote{https://github.com/marcoamonteiro/pi-GAN/blob/master/siren/siren.py\#L255}.\\

\noindent \textbf{2D Feature Grids}:
High-dimensional features stored in a high-resolution grid have proven to be effective in reducing rendering times for representing complex shapes~\cite{yu_and_fridovichkeil2021plenoxels,takikawa2021nglod, chen2022tensorf}. For example, given a 3D point $x\in \mathbb{R}^3$, we sample a feature from a learned high-resolution grid, e.g. $256^3$, by fetching $8$ nearest features in the grid and trilinearly interpolating in the cube formed by the $8$ neighboring features. We assume that the function we learn is linear in the high-dimensional feature space, and process the features using an MLP to obtain the value of the function at the given 3D point. Recent efforts~\cite{chen2022tensorf,chan2022eg3d} to factorize the 3D grids into three 2D grids have proven to be effective. We elucidate in the following text.\\ %

\noindent \textbf{2D Feature Grids for SDF:}

\noindent \textbf{For 3-dimensions:} A feature grid $M_{xyz} \in \mathbb{R}^{N \times N \times N} \times \mathbb{R}^K$ with resolution $N$ and $K$-dimensional features can be factored into three 2D feature grids $(M_{xy},M_{yz},M_{zx}) \in \mathbb{R}^{3 \times N \times N} \times \mathbb{R}^K$. Performing this factorization, we assume that the distribution of high-dimensional features in $(x,y)$ is independent of $z$, $(y,z)$ is independent of $x$, and that of $(z,x)$ is independent of $y$. We expect that the MLP handles cases where this assumption is broken. For SDF, we define three feature grids, ${(M^s_{xy},M^s_{yz},M^s_{zx}) \in \mathbb{R}^{3 \times N \times N} \times \mathbb{R}^K}$. Given a 3D point $x\in \mathbb{R}^3$, we retrieve the features ${m^s_{xy} \sim M^s_{xy}}$ \footnote{By `$\sim $' we mean to sample $4$ neigbouring features of a given 2D location in the grid and bilinearly interpolating between the features.}, ${m^s_{yz} \sim M^s_{yz}}$, and ${m^s_{zx} \sim M^s_{zx}}$, where $m^s_{xy}\,, m^s_{yz}\,, \text{ and } m^s_{zx} \in \mathbb{R}^K$. Using this, we define a function ${f^{ps}: \mathbb{R}^3 \rightarrow \mathbb{R}^{3 \times K}}$ as
\begin{equation}
    f^{ps}(x) = (m^s_{xy},m^s_{yz},m^s_{zx}). \label{eq:sdfgrid}
\end{equation}

\noindent \textbf{2D Feature Grids for DDF:} Our DDF representation is defined on a $6D$ grid, therefore we need to factorize a 6D grid into 2D grids. 
The number of 2D grids we need is $\binom{6}{2} = 15$, which is the number of 2D tuples we can make from a 6D tuple i.e., $(p_x,p_y,p_z,r_x,r_y,r_z)$ factorizes into $\{(p_x,p_y),(p_y,p_z)\,, \dots \,, (r_z,r_x)\}$ . For points $p = (p_x,p_y,p_z) \in \mathbb{S}^2$ on the unit sphere, and directions $r = (r_x,r_y,r_z) \in \mathbb{S}^2$, we define $15$ feature grids, $(M^d_{p_xp_y}, \dots, M^d_{r_zr_x}) \in \mathbb{R}^{15 \times N \times N} \times \mathbb{R}^K$.
Given a 6D tuple $(p,r) \in \mathbb{R}^6$, we retrieve the features ${m^d_{p_xp_y} \sim M^d_{p_xp_y}}\,, \dots\,, \text{ and } {m^d_{r_zr_x} \sim M^d_{r_zr_x}}$, where $m^d_{p_xp_y}\,,\dots\,, m^d_{r_zr_x} \in \mathbb{R}^K$. Using this, we define a function, ${f^{pd}: \mathbb{R}^6 \rightarrow \mathbb{R}^{15 \times K}}$ as
\begin{equation}
    f^{pd}(p,r) = (m_{p_xp_y}\,, \dots \,, m_{r_zr_x}). \label{eq:ddfgrid}
\end{equation}
Without this factorization, the memory required to store a 6D grid scales as $\mathcal{O}(N^6)$ for 2D featured grids, and is computationally highly inefficient. The 2D feature grids' memory requirements scale quadratically $\mathcal{O}(N^2)$ with the grid resolution. The 2D feature grids, $f^{pd}(p,r)$, and $f^{ps}(x)$, are learned per shape class and not per object. \\

\noindent \textbf{SDF Model:} The SDF model takes as input a latent code per shape $z \in \mathbb{R}^{L}$, a high-dimensional feature vector $f_{ps}(x) \in \mathbb{R}^{3\times K}$ from Eq.~\eqref{eq:sdfgrid}, and a point inside the unit sphere $x\in\mathbb{R}^3$. With that, it outputs an SDF value $s \in \mathbb{R}$. We learn the SDF model, $f_s: \{\mathbb{R}^{3},\mathbb{R}^{L}, \mathbb{R}^{3 \times K} \} \rightarrow \mathbb{R}$, using an MLP with parameters $\Theta_s$ as
\begin{equation}
    f_s(x,z,f_{ps}(x);\Theta_s) = s\,. \label{eq:sdfmodel}
\end{equation}
\noindent \textbf{DDF Model:} The DDF model takes as input a latent code (per shape) $z \in \mathbb{R}^{L}$, a high-dimensional feature vector $f_{pd}(x) \in \mathbb{R}^{15\times K}$ from Eq.~\eqref{eq:ddfgrid}, a tuple with a point on the unit sphere, and a direction $(p,r)\in\mathbb{R}^6$. The model outputs a DDF value $d \in \mathbb{R}_{+}$, and a ray hit probability $\sigma \in [0,1]$. We learn the DDF model ${f_d: \{\mathbb{R}^{6},\mathbb{R}^{L}, \mathbb{R}^{15 \times K} \} \rightarrow \{\mathbb{R}_{+},[0,1]\}}$ using an MLP with parameters $\Theta_d$ as
\begin{equation}
    f_d((p,r),z,f_{pd}(x);\Theta_d) = (d,\sigma)\,. \label{eq:ddfmodel}
\end{equation}
We encode the inputs to our models, $x$ and $(p,r)$, with positional encoding from NeRF~\cite{mildenhall2020nerf}.

\subsection{Training} \label{sec:trainingmethod}
We train a model for each class of the ShapeNet dataset~\cite{chang2015shapenet}.

\noindent \textbf{Data Preprocessing:} For training the network, we use the ground truth signed distance and the ground truth directional distance supervision. We use the preprocessing pipeline from DeepSDF~\cite{park2019deepsdf} to sample about $1$ million points for SDF supervision. We randomly sample $1$ million points on the unit sphere and random directions that point to the surface of an object using the object's point cloud for DDF supervision. We also sample $500k$ points and random missing directions. We render the $1.5$mil rays using Trimesh~\cite{haggerty2019trimesh} to obtain ground truth distances and ray hit supervision.\\

\noindent\textbf{Losses:}
We train the network with the following losses:
    \textbf{SDF loss} $\mathcal{L}_{s}$. We supervise the SDF network to predict signed distances $s$ (Eq.~\eqref{eq:sdfmodel}), with ground truth SDFs $s_{GT}$ using 
\begin{equation}
    \mathcal{L}_{s}(s) = \|s-s_{GT}\|_1 \,.
    \label{eq:sdfLoss}
\end{equation}
\textbf{DDF loss} $\mathcal{L}_{d}$. We learn the DDF model by supervising the model to predict directional distances $d$ (from Eq.~\eqref{eq:ddfmodel}) which are close to their corresponding ground truth distances $d_{GT}$ using
\begin{equation}
   \mathcal{L}_{d}(d) = \|d-d_{GT}\|_1 \,.
    \label{eq:ddfLoss}
\end{equation}
\textbf{Ray hit loss} $\mathcal{L}_{\sigma}$. We supervise the ray hit predictions $\sigma$ from the DDF model with the ray hit ground truths $\sigma_{GT}$ using the binary cross entropy loss as
\begin{equation}
   \mathcal{L}_{\sigma}(\sigma) = -(1-\sigma_{GT})\log(1-\sigma) - \sigma_{GT} \log(\sigma) \,.
    \label{eq:ddfrayhit}
\end{equation}
\textbf{TV regularizer} $\mathcal{L}_{tv}$. We enforce that the gradient of each of the 2D feature grids is small so that the features learned in the grid result in shapes that are not noisy, for both the feature grids of DDF and SDF models. The loss is given by
\begin{equation}
   \mathcal{L}_{tv}(M) = \sum_{i} \|\nabla M^s_{i}\|_2 + \sum_{i} \|\nabla M^d_{i}\|_2\,,
    \label{eq:tvlloss}
\end{equation}
where the gradients, $\nabla M^{s}_{i}$ and $\nabla M^{d}_{i}$, are computed using finite differences similar to how it is done for 3D feature grids~\cite{fang2022tineuvox}, ${i = {xy,yz,zx}}$ for SDF feature grids and ${i = {p_xp_y,p_yp_z,\dots,r_zr_x}}$ for DDF feature grids.\\

\noindent \textbf{Track-SDF Regularizer.} The predicted directional distance and the signed distance for an object need not agree, therefore, we additionally constrain that the DDF prediction results in a point close to the surface predicted by the SDF using the Track-SDF regularizer. Towards that, we compute the points using the predictions of the DDF model as $p+dr$ for point and direction pairs that point to the object surface. We enforce that these points are close to $0$ using
\begin{equation}
    \mathcal{L}_{ts}(d) =\|f_s(p+dr)\|_1 \,, \label{eq:tracksdf}
\end{equation}
where $(p, r)$ are point-direction tuples that point to a surface, and $d$ is the predicted directional distance for the point-direction tuples as in Eq.~\eqref{eq:ddfmodel}. Note that we only train the DDF model, and not the SDF model, with this loss.

\noindent \textbf{Latent code regularizer} $\mathcal{L}_{l}$. As we use an autodecoder framework~\cite{park2019deepsdf}, we enforce that the latent codes for different shapes are close to each other. This can be achieved by penalizing latent codes with large magnitudes so that latent codes are close to zero, i.e.,
\begin{equation}
   \mathcal{L}_{l}(z) = \|z\|_2 \,,
    \label{eq:latentRegularizer}
\end{equation}
where $z$ is the latent code for a given shape. Note that SDF and DDF have the same latent code for a given shape.
\textbf{Training loss}. The complete training loss is given as
\begin{equation}
\begin{split}
    \mathcal{L} &= w_{s}\mathcal{L}_{s} + w_{d}\mathcal{L}_{d} + w_{\sigma}\mathcal{L}_{\sigma} \\ 
    &+ w_{tv}\mathcal{L}_{tv} + w_{ts}\mathcal{L}_{ts} + w_{l}\mathcal{L}_{l}\,,
    \label{eq:totalLoss}
\end{split}
\end{equation}
where $w_{s}$, $w_{d}$, $w_{\sigma}$, $w_{tv}$, $w_{ts}$, and $w_{l}$ are the weights for the SDF loss, DDF loss, Ray hit loss, TV regularizer, Track-SDF regularizer, and latent code regularizer respectively. %
\noindent \textbf{Optimization:} We optimize the loss in Eq.~\eqref{eq:totalLoss} for the neural network weights, feature on the grid, and shape latent codes, $\Theta_s$, $\Theta_d$, $M$, and  $Z$, where $Z=\{z_i | i = 1\dots J\}$ is the set of latent codes representing all the $J$ training shapes, $\Theta_d$ are the learnable network parameters of the DDF model $f_d$, $\Theta_s$ are the learnable network parameters of the SDF model $f_s$, and $M = (M^s,M^d)$ are the SDF and DDF feature grids where ${M^s = \{M^s_{i} \,\, | \,\,  i = xy,yz,zx\}}$ are the SDF feature grids and ${M^d = \{M^d_{i} \,\, | \,\,  i = p_xp_y,\dots\,,r_zr_x\}}$ are the DDF feature grids.

\subsection{Reconstruction from Single-view Depth Maps} \label{sec:depthFitting}
Our autodecoder framework  allows us to work with any type of data without having to learn a new encoder for each type of data. Hence, during test time we merely need to optimize for the latent code $z$, while keeping the network and feature grids fixed.
The highlight of our reconstruction algorithm is that it obviates the need for sphere tracing at every iteration of the optimization.

For 3D reconstruction, we assume a depth map with an object mask and a given camera pose as input. We obtain the points of intersection of the rays $r$ from the camera with the unit sphere as $p$. At every iteration, we do the following:
\begin{enumerate}[leftmargin=*,wide, labelwidth=!, labelindent=0pt]
    \item with the latent code $z$ corresponding to the current iteration, evaluate the DDF model for the directional distance, $d,\sigma = f_d((p,r),f_{pd}(p,r),z)$ from $p$ along $r$ \label{it:step1}
    \item compute the 3D point inside the sphere predicted by the DDF model as $x=p+dr$ \label{it:step2}
    \item evaluate the SDF model at $x$ as $s= f_s(x,f_{ps}(x),z)$ \label{it:step3}
    \item optimize for the latent code $z$ of the object from the given depth map using the loss function, \label{it:step4}
\begin{equation}
    \mathcal{L}_{rec} = w_S\mathcal{L}_{S} + w_D\mathcal{L}_{D} + w_l\mathcal{L}_{l}\,,
    \label{eq:depthreconstruction}
    \end{equation}
    where $\mathcal{L}_{S}$ is the silhouette loss, $\mathcal{L}_{D}$ is the depth loss, and $\mathcal{L}_{l}$ is the regularizer (Eq.~\eqref{eq:latentRegularizer}) for learning the latent code with $w_S$, $w_D$, and $w_l$ as their respective weights. 
 Depth loss and silhouette loss are explained in the following.
\end{enumerate}%
\textbf{Depth Loss} $\mathcal{L}_D$. The depth loss is the error between the given depth $\lambda_{GT}$ and the predicted depth $\lambda$, i.e.
\begin{equation}
    \mathcal{L}_D(\lambda) = \| \lambda - \lambda_{GT}\|_1 \,. \label{eq:depthLoss}
\end{equation}
We obtain the predicted depth using $\lambda u = P x$, where $u$ are the image coordinates, $P$ is the given projection matrix of the camera and $x$ is the 3D point obtained in step~\ref{it:step2} above.
\textbf{Silhouette Loss}  $\mathcal{L}_{S}$. The silhouette loss is enforced as
\begin{align}
\mathcal{L}_{S}(s) &= \mathcal{L}_{S_{s_+}} + \mathcal{L}_{S_{s_-}} + \mathcal{L}_{\sigma}\,, \label{eq:depthsilhouettes}\\
 \mathcal{L}_{S_{s_+}} (m) &=\| \langle s,m \rangle \|_1\,, \label{eq:SDFPossilhouette}\\
 \mathcal{L}_{S_{s_-}}(m) &= \| |\langle s,1-m \rangle |-\tau \|_1  \,,\label{eq:SDFNegsilhouette}
\end{align}
where $m \in \{0,1\}$ is the given image mask, $s$ is the predicted signed distance from step~\ref{it:step3} above, $\tau$ is the truncation distance for the SDF model, $\sigma$ is the predicted ray hit probability from the DDF model in step~\ref{it:step1}, and $\mathcal{L}_{\sigma}$ is the DDF silhouette loss from Eq.~\eqref{eq:ddfrayhit}. The idea behind the loss is that where the rays hit the surface, the SDF must be as low as possible and where the rays don't, SDF must be high.

\section{Experiments}
In this section, we evaluate our method in different settings, reconstruction from single-view depth maps, and RGB videos. We evaluate the design choices of our method and the reconstruction algorithm in the ablation study. We show reconstruction from silhouettes and provide implementation details in supp. mat.
\subsection{Reconstruction from Single-view Depth Maps} \label{sec:3Drecfromdepthexp}
Our DDF model predicts distance to the surface of a shape given the latent code representing the shape, the ray origin, and the ray direction. Therefore, it can be used to replace the expensive sphere tracing algorithm during inverse rendering with learned SDF models.

We evaluate this advantage of our method by reconstructing the 3D shape given a depth map with a camera pose. We render a depth image with the given camera pose from our network and optimize for the latent code as discussed in Sec~\ref{sec:depthFitting}. 
We test our trained models on the first $200$ test instances of different classes of ShapeNet shapes -- airplanes, cars, chairs, lamps, sofas, and tables. For the images, we obtain the camera parameters of the first image of the rendered ShapeNet dataset from 3D-R2N2~\cite{choy20163d} and render a depth map with the same resolution, $137 \times 137$.
For comparisons, we run the official implementations of IF-Net~\cite{chibane2020ifnet} and DIST~\cite{liu2020dist}. With IF-Nets, we complete partial point clouds obtained by un-projecting the depth maps.\\
\begin{figure*}[ht]
    \centering
    
    \begin{overpic}[width=\linewidth,,tics=5]
        {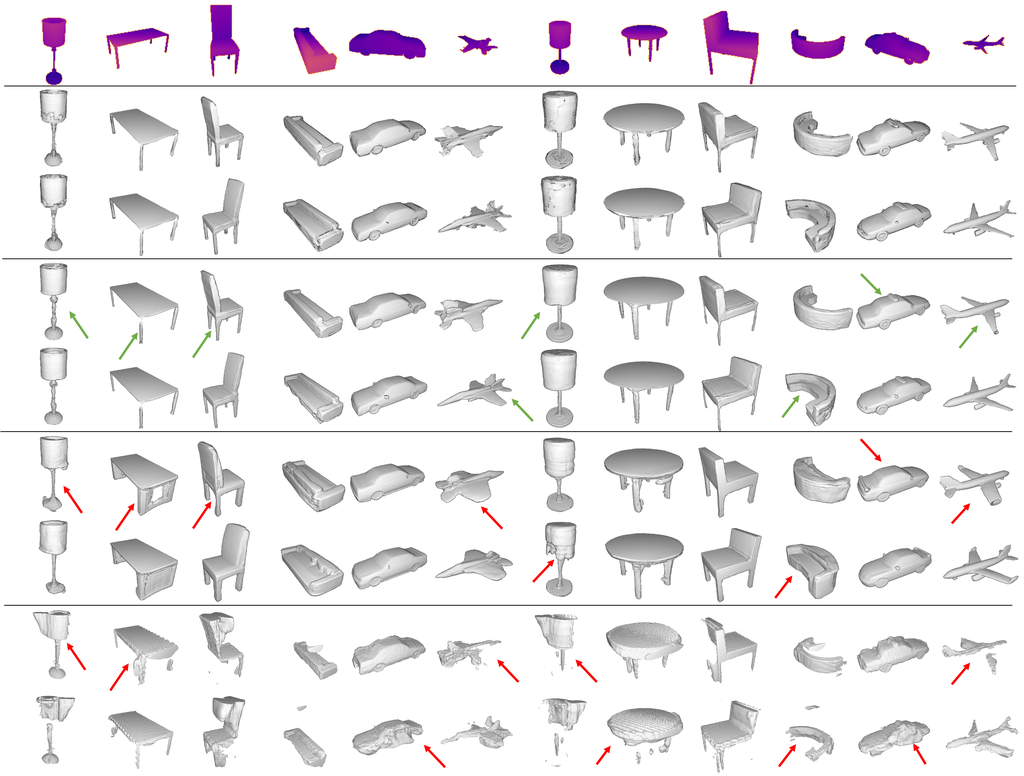}
        \put(0,68){\rotatebox{90}{\small Ground Truth}}
        \put(2,70.5){\rotatebox{90}{\small Depth}}
        
        \put(0,51.3){\rotatebox{90}{\small DDF Renders (Ours)}}
        \put(2,61){\rotatebox{90}{\small View1}}
        \put(2,52){\rotatebox{90}{\small View2}}

        \put(0,40){\rotatebox{90}{\small Ours}}
        \put(2,44){\rotatebox{90}{\small View1}}
        \put(2,35){\rotatebox{90}{\small View2}}

        \put(0,19){\rotatebox{90}{\small DIST + DeepSDF}}
        \put(2,27.5){\rotatebox{90}{\small View1}}
        \put(2,19){\rotatebox{90}{\small View2}}

        \put(0,6){\rotatebox{90}{\small IF-NET}}
        \put(2,10.5){\rotatebox{90}{\small View1}}
        \put(2,2){\rotatebox{90}{\small View2}}
    \end{overpic}
    \caption{3D shapes reconstructed from a given depth map. Each column shows reconstruction results for different shapes. Top row: Given depth map. Upper-middle rows: views rendered with $1$ forward pass from our DDF model. Middle rows: views of 3D shapes reconstructed by our SDF model. Lower-middle rows: views of 3D shape reconstructed using DIST~\cite{liu2020dist}. Last rows: views of 3D shapes reconstructed using IF-NET~\cite{chibane2020ifnet}. Our method outperforms existing methods, as it can, for example, better model fine-scale details (see e.g.~the legs of the tables or chairs, or the geometry of the airplanes)}
    \label{fig:comparison}
    \vspace{-0.5cm}
\end{figure*}
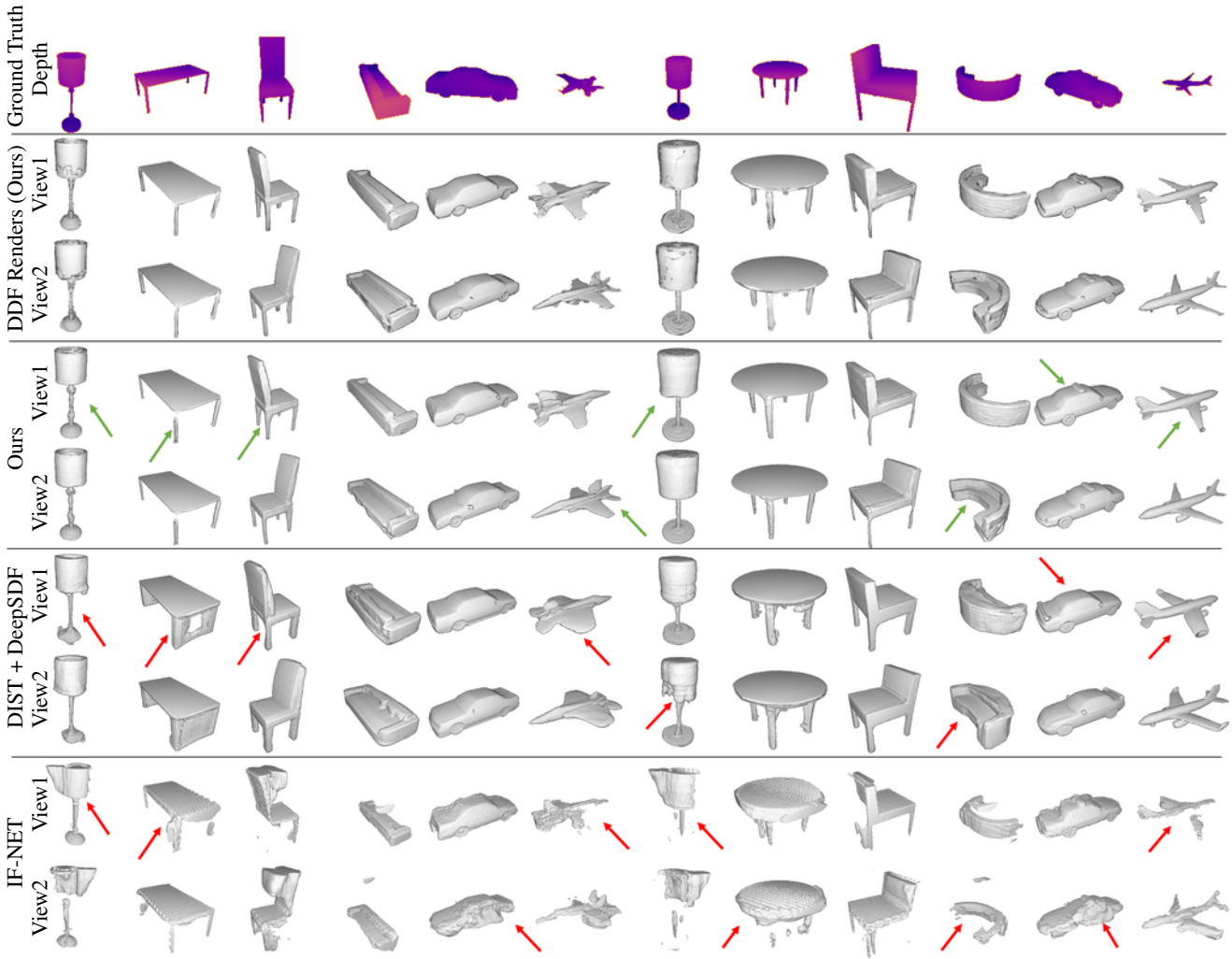
\noindent\textbf{Qualitative Results:} We show qualitative results in Fig.~\ref{fig:comparison}. 
It can be seen that our method can reconstruct 3D shapes accurately given a single view depth image. Given our feature-based network architecture, and our algorithm, our reconstructions are more detailed compared with DIST~\cite{liu2020dist}. Further, our method is about $15.5\times$ faster per iteration on average (see Tab.~\ref{table:comparisons}). We compare our reconstruction with those of IF-NET~\cite{chibane2020ifnet}, a state-of-the-art encoder-based neural implicit representation. While IF-NET leads to plausible reconstructions in the observed locations, where there are valid depth maps, it does not complete unobserved shapes, as shown in the last two rows of Fig.~\ref{fig:comparison}.\\
\begin{table*}[ht]
    \centering
    \begin{tabular}{l||c|c|c|c|c||c|c||c|c|c}
         \textbf{Method} &  \textbf{Ours} &  \textbf{Ours} & \multicolumn{2}{c|}{\textbf{DIST}} &  \textbf{IF-NET} &  \textbf{Ours} &  \textbf{DIST} & \textbf{Ours} & \textbf{Ours} & \textbf{DeepSDF} \\
         & \textbf{SDF} & \textbf{DDF} & \textbf{Our SDF} & \textbf{DeepSDF} & &  &  \textbf{DeepSDF} & \textbf{DDF} & & \\ \hline
         \textbf{Metric} & \multicolumn{5}{c||}{$1000\times$ CD $\downarrow$} & \multicolumn{2}{c||}{ms/iteration $\downarrow$} & \multicolumn{3}{c}{ms / $256\times256$ frame $\downarrow$} \\ \hline
         Car   & 0.55 & \textbf{0.38} & 0.60 & 0.61 & 4.09 & \textbf{17} & 282 & 62 & \textbf{23} & 132 \\ 
         Chair & 0.74 & \textbf{0.64} & 1.96 & 1.92 & 5.45 & \textbf{18} & 236 & 58 & \textbf{23} & 118\\ 
         Lamp  & \textbf{2.50} & 4.81 & 6.39 & 7.34 & 6.05 & \textbf{15} & 281 & 65 & \textbf{22} & 120\\ 
         Plane & \textbf{0.18} & 0.32 & 0.69 & 0.94 & 2.08 & \textbf{15} & 231 & 56 & \textbf{22} & 116\\ 
         Sofa  & 0.77 & \textbf{0.67} & 1.64 & 1.81 & 9.43 & \textbf{18} & 238 & 61 & \textbf{21} & 113\\  
         Table & 1.28 & \textbf{0.83} & 3.02 & 2.79 & 4.67 & \textbf{18} & 283 & 56 & \textbf{23} & 126\\
    \end{tabular}
    \caption{Quantitative results of comparisons of our method with DIST~\cite{liu2020dist} and IF-NET~\cite{chibane2020ifnet} (middle-left columns). Our method outperforms DIST and IF-NET in all the shape classes, showing that our models and our depth-fitting algorithm lead to better reconstructions. DIST performs marginally better in most classes with our SDF model compared with DeepSDF's, showing that the majority of improvement is due to our method and not the SDF model. We compare the time per optimization step with DIST, where ours is on an average $15.5\times$ faster than DIST, as shown in the middle-right columns. Finally, in the right-most columns, we show that we can render $256\times 256$ images in real-time with just one forward pass using our DDF representation. Rendering times include time for normal computation.
    }
    \label{table:comparisons}
\end{table*}
\noindent\textbf{Quantitative Results:} We show the quantitative results in Tab.~\ref{table:comparisons}. We use the chamfer distance defined in DeepSDF~\cite{park2019deepsdf} to compute the accuracy. Please see supp. mat. for more details. The results are consistent with qualitative ones, as our method can fit well to the given depth maps and obtain more plausible reconstructions compared to DIST~\cite{liu2020dist}. Moreover, we outperform DIST in all the classes while being $15.5\times$ faster. Further, as IF-NET~\cite{chibane2020ifnet} does not complete the shape in unobserved areas, we significantly outperform IF-NET quantitatively and qualitatively.

\textbf{Model Evaluation:} We compare our model with the state-of-the-art directional distance representation methods, PRIF~\cite{feng2022prif}, Depth-LFN~\cite{sitzmann2021lfns}, and NeuralODF~\cite{houchens2022neuralodf} on reconstruction from depth maps. PRIF predicts the directional distance from the perpendicular foot of a camera ray. LFN predicts RGB given Plücker coordinates and ray direction. NeuralODF predicts distance to the surface, and ray hit prediction, given a point and direction in 3D space. As predicting DDF everywhere in space is a harder task, for a fair comparison, we restrict the input to NeuralODF to the unit sphere. Further, we found that predicting ray hit from the final layer results in higher accuracy for NeuralODF; hence, we use this model. For a fair comparison, we train LFNs with just the depth and ray hit supervision so that the model can predict dense depth. We train the three models on the first $256$ shapes of the training set and test them on the first $64$ shapes of the test set, of each shape class. We follow this split to closely replicate the number of train and test shapes in PRIF. We evaluate the methods quantitatively using chamfer distance between the predicted and ground truth shapes.

We optimize for latent code using the algorithm in Sec.~\ref{sec:3Drecfromdepthexp} during inference with our method. For other methods, we optimize for the latent code with the losses from Eqs.~\eqref{eq:depthLoss} and ~\eqref{eq:ddfrayhit}. Our method marries the best of both the models, view-consistent geometric details from the SDF model and $1$ forward pass rendering from the DDF model. Owing to this, our model outperforms the state-of-the-art DDF models by a large margin quantitatively, as seen in Tab.~\ref{tab:ddfcomp}, and qualitatively (see supp. mat. Fig. 3).

\begin{table}[t]
    \centering
    \begin{tabular}{l|cc|c|c|c}
                    & \multicolumn{2}{c|}{\textbf{Ours}}  &   \textbf{PRIF} &   \textbf{LFN} &   \textbf{Neural} \\
   \textbf{Class}     & \textbf{SDF}    &\textbf{DDF}   &  &  & \textbf{ODF} \\ \hline
            & \multicolumn{5}{c}{$1000\times$ CD $\downarrow$ (Mean)}\\ \hline
            Cars       & 0.71   & \textbf{0.57}  &  0.85    & 0.66   & 0.83 \\ 
            Chairs     & 1.30   & \textbf{1.15}  &  1.83    & 1.56   & 1.78 \\ 
            Lamps    & \textbf{4.98}   & 6.52    &  9.22    & DNC    & 7.68 \\ 
            Planes  & \textbf{0.26}   & 0.51     &  0.78    & 0.59   & 0.68 \\ 
            Sofas      & 0.78   & \textbf{0.78}  &  1.57    & 1.08   & 1.57 \\ 
            Tables     & 1.49   & \textbf{1.40}  &  2.30    & 1.63   & 2.60 \\ 
    \end{tabular}
    \caption{
    Quantitative comparison of our model with PRIF~\cite{feng2022prif}, Depth-LFNs~\cite{sitzmann2021lfns}, and NeuralODF~\cite{houchens2022neuralodf} on reconstruction from depth maps. We train the models on different classes of shapes and utilize them in the autodecoder framework to optimize for shapes from a given depth map during inference. We report the mean chamfer distance between the reconstructed and ground truth shapes. Our model outperforms competitive DDF models in all the classes. While our model maintains the salient features of DDF, such as $1$ forward pass rendering, it can also represent view-consistent geometric details using the SDF model. (Depth-LFN did not converge for lamps class with $256$ shapes.)
    }
    \label{tab:ddfcomp}
    \vspace{-5mm}
\end{table}

\begin{table*}[ht]
    \centering
    \setlength\tabcolsep{2pt}
    \begin{tabular}{l|cccccc|c|cccccc}
        & \multicolumn{6}{c|}{\textbf{Reconstruction Algorithm}} & & \multicolumn{6}{c}{\textbf{Model and Losses}}  \\ \hline
          &   \textbf{DIST}   & \textbf{wo $\mathcal{L}_{S}$}  & \textbf{wo $\mathcal{L}_{S_{s_+}}$ }    & \textbf{wo $\mathcal{L}_{S_{s_-}}$ }                        & \textbf{wo $\mathcal{L}_{S_{s_+}}$}                                                                    
         
         & \textbf{wo $\mathcal{L}_{\sigma}$} & \textbf{Ours} &  \textbf{wo sh.} & \textbf{w $\mathcal{L}_{ts}$}          & \textbf{wo $\mathcal{L}_{ts}$}    & \textbf{wo $\mathcal{L}_{tv}$}    & \textbf{wo DDF}     &  \textbf{w SDF} \\

           &       & \textbf{Eq.~\eqref{eq:depthsilhouettes}}  & \textbf{Eq.~\eqref{eq:SDFPossilhouette}} & \textbf{Eq.~\eqref{eq:SDFNegsilhouette}} &$+ \mathcal{L}_{S_{s_-}}$  & \textbf{Eq.~\eqref{eq:ddfrayhit}}    & & \textbf{lats.}    & \textbf{to SDF}  & \textbf{Eq.~\eqref{eq:tracksdf}} & \textbf{Eq.~\eqref{eq:tvlloss}}   & \textbf{$\sigma$ preds.}    & \textbf{3D Grid} \\ \hline
                      
         SDF          & 1.56 & 2.74 & 7.92 & 1.38 & 2.62 & 0.96 & {0.78} & 0.98 & 1.48 & 0.93 & 0.77 & 0.96 & 0.87 \\
         DDF          & 1.72 & 1.37 & 0.89 & 1.13 & 1.00 & 1.00 & {0.78} & 0.95 & 1.49 & 0.79 & 1.26 & 1.00 & 0.84 \\
    \end{tabular}
    \caption{
    Quantitative results of ablation study. Ablations on reconstruction algorithm (left columns), and ablations on models (right columns). \textit{Reconstruction algorithm}: from left to right, DIST algorithm with our model, without any silhouette losses $\mathcal{L}_S$, without foreground SDF silhouette loss $\mathcal{L}_{S_{s_+}}$, without background SDF silhouette loss $\mathcal{L}_{S_{s_-}}$, without any SDF silhouette losses, without the DDF silhouette loss $\mathcal{L}_\sigma$, and ours. \textit{Model}: From left to right: without a shared latent space for SDF and DDF models, with gradients from Track-SDF regularizer $\mathcal{L}_{ts}$ to SDF model, without Track-SDF regularizer $\mathcal{L}_{ts}$, without the TV regularizer $\mathcal{L}_{tv}$, without DDF ray hit predictions $\sigma$, and with a 3D feature grid instead of a 2D grid for SDF model. Middle: Our proposed algorithm.
    }
    \label{tab:ablations}
\end{table*}

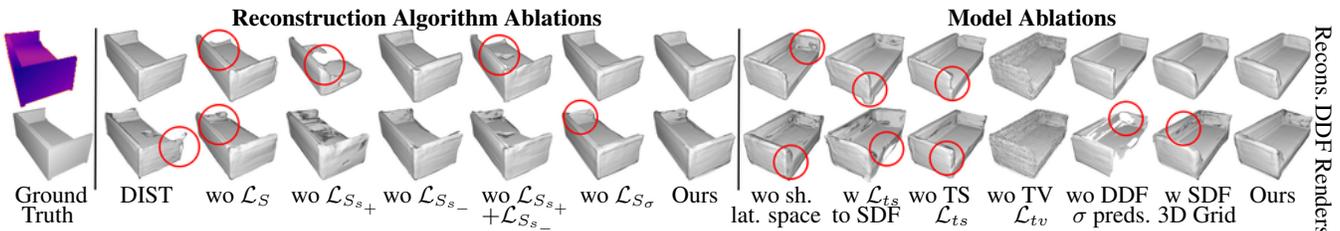
\begin{figure*}[t]
    \centering
    \vspace{3mm}
    \begin{overpic}[width=\linewidth,,tics=2]
        {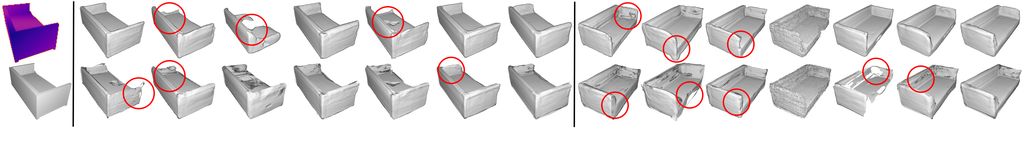}
        \put(100,14){\rotatebox{-90}{\small Recons.}}
        \put(100,8){\rotatebox{-90}{\small DDF Renders}}
        \put(17.5,14){{\small \textbf{Reconstruction Algorithm Ablations}}}
        \put(72,14){{\small \textbf{Model Ablations}}}
        \put(1,    0.5){{\small Ground}}
        \put(1.5, -1){{\small Truth}}
        
        \put(9,  0.5){{\small DIST}}
        
        \put(15.5,0.5){{\small wo $\mathcal{L}_{S}$}}
        
        \put(22,0.5){{\small wo $\mathcal{L}_{S_{s_+}}$}}
        
        \put(29,0.5){{\small wo $\mathcal{L}_{S_{s_-}}$}}
        
        \put(36.5,0.5){{\small wo $\mathcal{L}_{S_{s_+}}$}}
        \put(36.5,-1){{\small $+\mathcal{L}_{S_{s_-}}$}}

        \put(44,0.5){{\small wo $\mathcal{L}_{S_{\sigma}}$}}

        \put(51,0.5){{\small Ours}}

        \put(57,0.5){{\small wo sh.}}
        \put(55.5,-1){{\small lat. space}}

        \put(64,0.5){{\small w $\mathcal{L}_{ts}$}}
        \put(63.2,-1){{\small to SDF}}
        
        \put(69,0.5){{\small wo TS}}
        \put(71,-1){{\small $\mathcal{L}_{ts}$}}
        
        \put(75,0.5){{\small wo TV}}
        \put(77,-1){{\small $\mathcal{L}_{tv}$}}
        
        \put(81,0.5){{\small wo DDF}}
        \put(81.5,-1){{\small $\sigma$ preds.}}
        
        \put(88.5,0.5){{\small w SDF}}
        \put(88,-1){{\small 3D Grid}}

        \put(95,0.5){{\small Ours}}
    \end{overpic}
    \caption{
    Qualitative results of ablation study. Left: Ground truth depth (top) and geometry (bottom). Middle: \textit{Reconstruction algorithm}: (left to right) with DIST and our SDF, without any silhouette loss $\mathcal{L}_S$, without foreground SDF silhouette loss $\mathcal{L}_{S_{s_+}}$, without background SDF silhouette loss $\mathcal{L}_{S_{s_-}}$, without any SDF silhouette loss $\mathcal{L}_{S_{s_+}}+\mathcal{L}_{S_{s_-}}$, without DDF silhouette loss $\mathcal{L}_\sigma$, and ours. Right: \textit{Model}: (left to right) without a shared latent space between DDF and SDF, track SDF $\mathcal{L}_{ts}$ loss also trains SDF model, without track SDF $\mathcal{L}_{ts}$, without TV regularizer $\mathcal{L}_{tv}$, without DDF ray hit predictions $\sigma$, with a 3D feature grid for SDF instead of 2D grids, and ours. Our design choices result in fast and accurate reconstructions. 
    }    \vspace{-0.5cm}

    \label{fig:ablations}
\end{figure*}
\subsection{Ablations} \label{sec:ablations}
In this experiment, we evaluate the design choices in our method. We train the models with the first $256$ shapes from the training split and test on the first $64$ test shapes of the sofas class from the ShapeNet dataset. We report $1000\times$ the chamfers distance between reconstructions and ground truth in Tab.~\ref{tab:ablations}.\\
\textbf{Reconstruction Algorithm:} We train our models as described in Sec.~\ref{sec:trainingmethod}. We reconstruct 3D shapes from depth maps (see Sec.~\ref{sec:3Drecfromdepthexp}) using our learned models. We ablate the components of losses introduced in Sec.~\ref{sec:depthFitting}. Quantitative results are shown in Tab.~\ref{tab:ablations} and qualitative results are shown in Fig.~\ref{fig:ablations}. 
\textit{(DIST)} We run the single view reconstruction algorithm with DIST on our trained model. As DDF and SDF share a latent space, we can also evaluate DDF. Quantitatively DIST underperforms as we have also seen in Tab.~\ref{table:comparisons}.
\textit{(wo $\mathcal{L}_S$)} Without any silhouette losses, our method performs poorly, showing the impact of silhouette losses on reconstruction quality. 
\textit{(wo $\mathcal{L}_{s_+}$)} Without the foreground SDF silhouette loss, the background silhouette loss overpowers the reconstruction and leads to missing regions, as shown in Fig.~\ref{fig:ablations}, $4^{\text{th}}$ column.
\textit{(wo $\mathcal{L}_{s_-}$)} Without a background silhouette loss, the SDF reconstruction can be larger than the masks, leading to poor accuracy. 
\textit{(wo $\mathcal{L}_{s_+}$+ $\mathcal{L}_{s_-}$)} Without any SDF silhouette losses, the SDF reconstructions miss structures leading to poor accuracy. \textit{(wo $\mathcal{L_\sigma}$)} Without the DDF silhouette loss the DDF renders are inconsistent with SDF reconstructions. Since we use DDF for rendering the SDF, this also leads to a decrease in the reconstruction quality of the SDF. \\
\textbf{Model:}
We perform ablation studies on our models and losses presented in Sec~\ref{sec:method}. Qualitative results are shown in Fig.~\ref{fig:ablations}, and quantitative results are shown in Tab.~\ref{tab:ablations}. 
\textit{(wo shared latent space)} Independent optimization for latent codes does not let the DDF and SDF models change together, leading to inaccuracies between the reconstructions as shown in Fig.~\ref{fig:ablations}. 
\textit{(with $\mathcal{L}_{ts}$ to SDF)} When the SDF model is allowed to train with the gradients from the Track-SDF regularizer (Eq.~\eqref{eq:tracksdf}), the reconstructions are bad as the SDF model can incorrectly learn to place a surface at DDF's predictions during training.
\textit{(wo TS $\mathcal{L}_{ts}$)} As the DDF is unconstrained the accuracy increases, however, since the DDF model does not predict close to the SDF surface, the reconstruction quality of SDFs is lower.
\textit{(wo TV $\mathcal{L}_{tv}$)} Without the TV regularizer, the reconstructions are noisy around the surface but sharp for SDF hence leading to high accuracy whereas for DDF this noise leads to higher error as the predicted distances are noisy. 
\textit{(wo DDF $\sigma$ preds.)} without DDF ray hit predictions, the DDF renders are incomplete, as we rely on SDF value for the ray hit predictions, leading to poor accuracy. 
\textit{(w 3D Grid SDF)} As a 3D feature grid with the same number of parameters as a 2D feature grid is of lower resolution, the model performs worse with the same number of features in the grid, leading to smoother reconstructions. Our model performs the best with all the design choices. 
\section{Future Work}
Overall, our method has shown significant improvement in terms of speed while reconstructing from different inputs such as depth maps, silhouettes (supp. mat.), and videos. 
While our results show consistent renders from different views using the DDF model, 3D inconsistency is a persistent problem with directional representations. The feature grid-based representation achieves high-quality results, however, better regularizers than the TV (Eq.~\ref{eq:tvlloss}) that allow for discontinuities while suppressing noise could help improve the representational capacity of DDF models.

\section{Conclusion}
We presented a novel 3D representation, DDF, that enables us to replace sphere tracing for rendering SDFs with just $1$ network evaluation per camera ray. Based on the learned DDF and SDF models, we introduced a fast algorithm (FIRe) to reconstruct shapes with our learned models from depth maps. We experimentally showed that FIRe can reconstruct high-quality 3D shapes given a depth map or a video while achieving an order of magnitude speedup of the optimization algorithm. We believe that the proposed method can play a crucial role in working with learned implicit scene representations for various applications. In order to stimulate follow-up work we plan to make our code publicly available.

\section{Acknowledgements} \vspace{-1.5mm}
{\small This work has been supported by the ERC
Advanced Grant SIMULACRON (884679), and the BMBF-funded Munich Center for Machine Learning. Thanks to Mariia Gladkova for helping set up the camera pipelines.}

\clearpage
\appendix
In this supplementary material, we provide the implementation details, such as our MLP architecture, and hyperparameters. We show additional qualitative and quantitative results. Finally, we evaluate our model for 3D shape reconstruction from single-view silhouettes and compare the results with DIST's~\cite{liu2020dist}.

\section{Implementation Details}
In this section, we provide details about our training, inference, network architecture, and hyperparameters.

\textbf{Training:} We train a model for each of the $6$ classes of the ShapeNet dataset with the splits from DeepSDF~\cite{park2019deepsdf}. We use a batch size of $64$ and $4096$ samples per scene. We train for $3000$ iterations for classes with less than $3000$ shapes, and $2000$ iterations otherwise. Each batch takes about $2.5$s on an Nvidia A100 GPU. Depending on the number of shapes in a class it takes $1-4$ days to train a model per class. We do not train the SDF model with the Track-SDF regularizer in Eq. 10 (main). We truncate the SDF values for better shape representations~\cite{park2019deepsdf}.

\textbf{Inference:} We run all the inference tasks on an Nvidia A40 GPU both for ours and DIST~\cite{liu2020dist} to ensure that the optimization and rendering performance we report are consistent.

\subsection{Hyperparameters}

We optimize the losses during both training and inference with ADAM~\cite{kingma2014adam} algorithm in Pytorch~\cite{paszke2019pytorch}. During \textbf{training}, we set the weights $w_{s}=1.0$, $w_{d}=1.0$, $w_{\sigma}=1.0$, $w_{tv}=100.0$, $w_{ts}=0.1$, and $w_{l}=0.0001$ respectively for SDF loss, DDF loss, Ray hit loss, TV regularizer, Track-SDF regularizer, and latent code regularizer. We train our models with a learning rate of $0.0005$ and our latent codes with a learning rate of $0.001$. After every $750$ or $500$  iterations (for $3000$ or $2000$ total iterations, respectively), we divide the learning rate by $2$. During \textbf{inference}, for 3D reconstruction from single-view depth maps, we set $w_S=1.0$, $w_D=1.0$, and $w_l=0.0001$ for the silhouette loss, the depth loss, and the latent regularizer respectively. For 3D reconstruction from single-view silhouettes, we set $w_S=1.0$, $w_D=0.0$, and $w_l=0.005$. We run our algorithm for $1000$ iterations with a starting step size of $0.001$ and a step size of $0.0005$ after $500$ iterations. We set a threshold of $0.8$ for DDF ray hit prediction for rendering. We extract meshes at the level-set of SDF, $s=0.001$. Similar to DeepSDF~\cite{park2019deepsdf}, we set a truncation distance $\tau = 0.1$. In other words, SDF values that are more than $0.1$ are set to $0.1$ and those less than $-0.1$ are set to $-0.1$.  We use the released code and parameters of DIST~\cite{liu2020dist} and IF-NET~\cite{chibane2020ifnet} for comparisons.
\subsection{Network Architecture}

\noindent \textbf{DDF Model ($f_d$):} \\
\textbf{2D Feature Grids:} For each of the $15$ feature grids, we set a resolution of $512\times 512$ and a feature dimension of $32$. \textbf{MLP:} The MLP in the DDF model consists of $3$ blocks of fully-connected layers. Each fully-connected block has $1$ hidden layer of $512$ dimension. There are skip connections to each block from the input. We use ReLU activations for the outputs of all the layers except the last layer where we use no activation function. We positionally encode~\cite{mildenhall2020nerf} each dimension of the input points and direction tuple with $3$ frequencies.

\noindent \textbf{SDF Model ($f_s$):} \\
\textbf{2D Feature Grids:} For each of the $3$ feature grids, we set a resolution of $512\times 512$ and a feature dimension of $32$. \textbf{MLP:} The MLP in the SDF model consists of a fully-connected block with $2$ hidden layers of $256$ dimension. We use ReLU activations for the outputs of all the layers except the last layer where we use no activation function. We positionally encode~\cite{mildenhall2020nerf} each dimension of the input points with $3$ frequencies.

\subsection{Quantitative Evaluation Metrics} \label{sec:quantitativeEval}
We extract meshes at a resolution of  $256^3$ using marching cubes to evaluate our SDF reconstructions quantitatively. To evaluate our DDF reconstructions, we first randomly sample points on the unit sphere and directions. We use rejection sampling by means of the ray hit predictions from our DDF model to obtain point-direction pairs that hit the object's surface. We obtain a point for each point-direction pair using $o+dr$ where $(o,r)$ is the point-direction pair that points to the surface and $d$ is the predicted directional distance.
We use the symmetric L2 chamfers distance as the metric for our quantitative experiments, i.e.
\begin{equation}
    \text{CD} = \frac{1}{N}\Sigma_{i=1}^N \min_{y\in Y} \|x_i-y\|^2_2 +  \frac{1}{N} \Sigma_{i=1}^N \min_{x\in X} \|x-y_i\|^2_2 \,,
    \label{eq:chamfer}
\end{equation}
where ${X = \{ x_i\in \mathbb{R}^3 |\, i=1\,,\dots\,,N \}}$ and ${Y = \{ y_i\in \mathbb{R}^3 |\, i=1\,,\dots\,,N \}}$ are points on the surfaces of two shapes, and $N$ is the number of points sampled on the two shapes. We compute chamfer's distance between $30000$ points from the two sets in the ShapeNet dataset's scale similar to DeepSDF~\cite{park2019deepsdf}.

\section{Experiments}
In this section, we show additional qualitative results on 3D shape reconstruction from depth maps in Fig.~\ref{fig:comparisonSupp}. We show qualitative results on our DDF model evaluations of the state-of-the-art directional distance models in Fig.~\ref{fig:prifcomparison}. We show results on reconstructing shapes from synthetic and real-world videos using our trained model in Sec.~\ref{sec:recvideo}, followed by shape reconstruction from silhouette experiments in Sec.~\ref{sec:silRec}. Finally, we provide details about our visualizations in Sec.~\ref{sec:visuals}.

\begin{table}[h]
    \centering
    \begin{tabular}{l|ccc|cc}
                        & \multicolumn{3}{c|}{\textbf{Ours}}  &   \multicolumn{2}{c}{\textbf{DIST}} \\ \hline
               \textbf{Class}    & \textbf{SDF}    & \textbf{DDF}   & \textbf{Time (s)} &      \textbf{DeepSDF}   & \textbf{Time (s)} \\\hline
                Cars    & 0.62   & 0.48  &  2.59  &  0.69 &  29.54          \\
                Planes  & 1.60   & 1.58  &  2.59  &  1.44 &  30.18          \\
                Sofas   & 1.80   & 1.87  &  2.50  &  1.82 &  29.89          \\
    \end{tabular}
    \caption{
    Quantitative results ($1000\times$ CD) of 3D reconstruction from video sequences in PMO~\cite{lin2019photometric}. Our method is on average $~12 \times$ faster than DIST~\cite{liu2020dist} while being as accurate.
    }
    \label{tab:rgbrec}
    \vspace{-0.5cm}
\end{table}
\subsection{Reconstruction from Videos} \label{sec:recvideo}
\begin{figure*}[t]
    \centering
    \begin{overpic}[width=\linewidth,,tics=2]
        {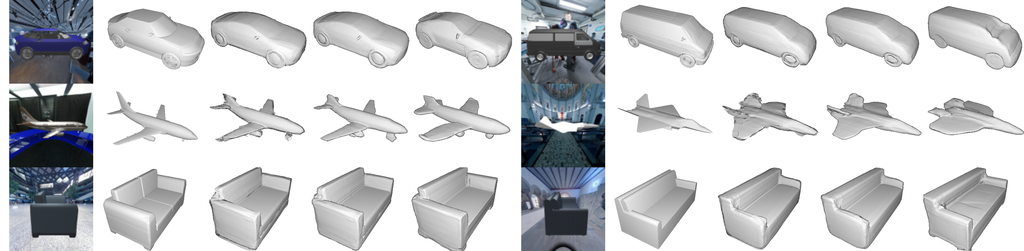}
        \put(3,-3){\small Video}
        \put(11,-3){\small Ref. GT}
        \put(20,-3){\small Ours (DDF)}
        \put(30,-3){\small Ours (SDF)}
        \put(43,-3){\small DIST}
        \put(53,-3){\small Video}
        \put(61,-3){\small Ref. GT}
        \put(70,-3){\small Ours (DDF)}
        \put(80,-3){\small Ours (SDF)}
        \put(93,-3){\small DIST}
    \end{overpic}
    \vspace{2mm}
    \caption{Qualitative results of 3D reconstruction from videos of PMO~\cite{lin2019photometric}. In both columns, Left to right: a frame from the input video, reference ground truth geometry, $1$ forward pass renders from our DDF, our SDF reconstructions, and reconstructions with vanilla DIST's algorithm. When we replace the raymarching algorithm in the vanilla implementation (DIST~\cite{liu2020dist}) with our DDF model, the algorithm is accelerated by about $12\times$ as our model only needs $1$ forward pass through the network per ray to render.}
    \label{fig:rgbrec}
\end{figure*}

While our representation performs well with the algorithm that we propose, we evaluate its capacity to accelerate existing algorithms. Towards that, we replace the sphere tracer in DIST's reconstruction from videos of PMO~\cite{lin2019photometric} dataset. We are given camera poses for the frames without object masks. We optimize for the latent code as
\begin{align*}
    \underset{z}{\operatorname{argmin}} \sum_{i=0}^{N-1} \sum_{j\in\mathcal{N}_i} \| I_i - I_{j \rightarrow i} (f(z)) \| + \mathcal{L}_{s_+}(\sigma) + \mathcal{L}_{s_-}(\sigma)\,,
\end{align*}
where $N=72$ is the number of frames of the video, $I_i$ is the $i^{\text{th}}$ image and $I_{j \rightarrow i}$ is an image formed using pixel colors warped from image $j$ to $i$ using the rendered depth, $f(z)$, obtained from DDF model as $ (f(z), \sigma) = f_d((p,r),z,f_{pd};\Theta_d)$ (see Eq. (5), main), $\sigma \in \{0,1\}$ is the predicted mask, and $\mathcal{N}_i$ is the neighborhood of frame $i$. We enforce the constraints $\mathcal{L}_{S_{s_+}}(\sigma)$ and $\mathcal{L}_{S_{s_-}}(\sigma)$, from Eqs. (16) and (17) (main) using predicted masks $(\sigma)$ to ensure that the predicted SDF surface follows DDF.
We show the qualitative results in Fig.~\ref{fig:rgbrec}. We show quantitative results in Tab.~\ref{tab:rgbrec}. As corroborated in the ablations (wo $\mathcal{L}_S$), our algorithm performs similarly compared to DIST when ground truth masks are unavailable. However, as we do not need to perform the expensive sphere tracing in each iteration, our method is $12\times$ faster. This shows the potential of our trained DDF model to replace sphere tracing in existing algorithms for an order of magnitude acceleration.

\begin{figure}[t]
    \centering
    \begin{overpic}[width=\linewidth,,tics=5]
        {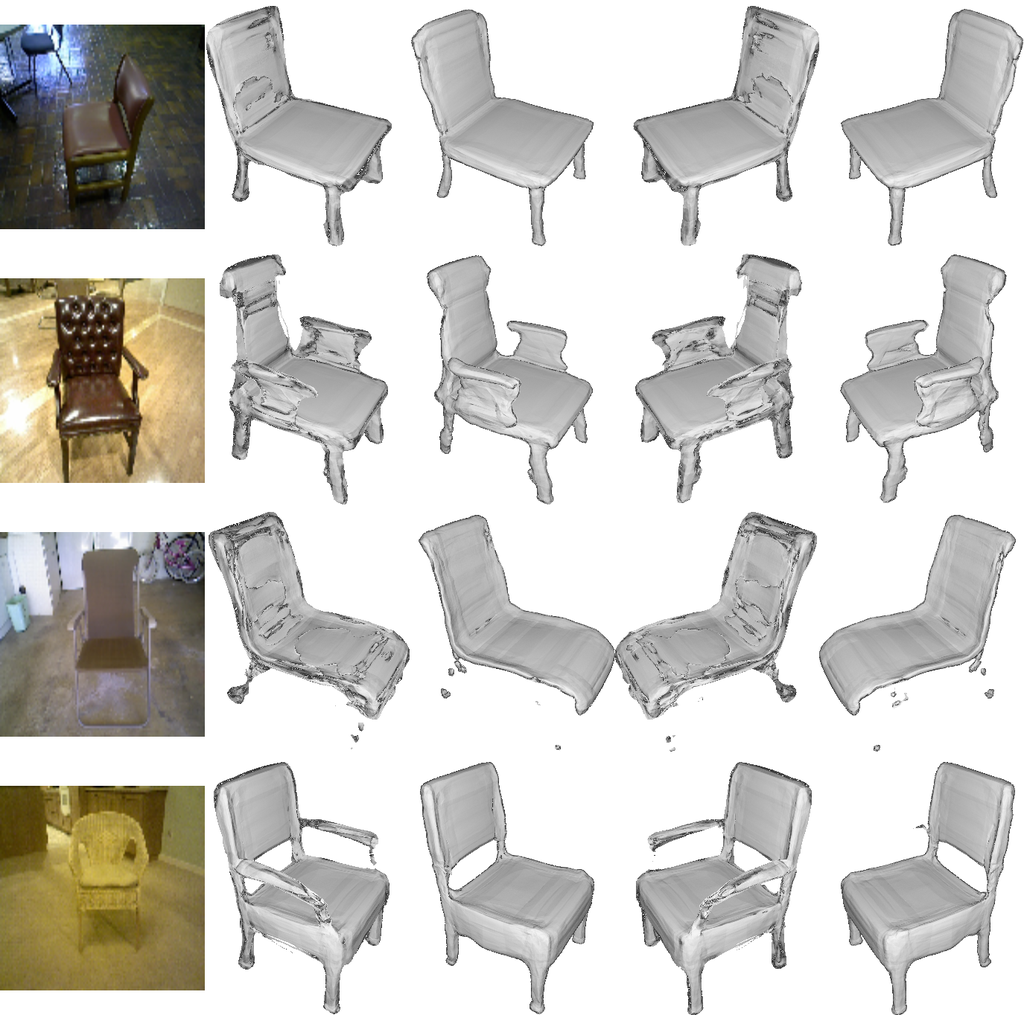}
        \put(7 ,-4){\small Video}
        \put(20,-4){\small Ours (DDF)}
        \put(40,-4){\small Ours (SDF)}
        \put(60,-4){\small Ours (DDF)}
        \put(80,-4){\small Ours (SDF)}
        \put(33,100){\small View 1}
        \put(78,100){\small View 2}
    \end{overpic}
    \vspace{3mm}
    \caption{Qualitative results of 3D reconstruction from $40$ real-world videos of chairs from Redwood dataset~\cite{choi2015redwood}. We reconstruct chairs by optimizing for photometric consistency between different frames of videos using the depth predicted by our model. Left: A frame from the input video. Right: DDF renders and reconstructed SDF in two views. The average distance from the predicted shape to the ground truth mesh is $4.0$cm.}
    \label{fig:mvQualResReal}
\end{figure}

\begin{figure*}[t]
    \centering
    
    \begin{overpic}[width=\linewidth,,tics=5]
        {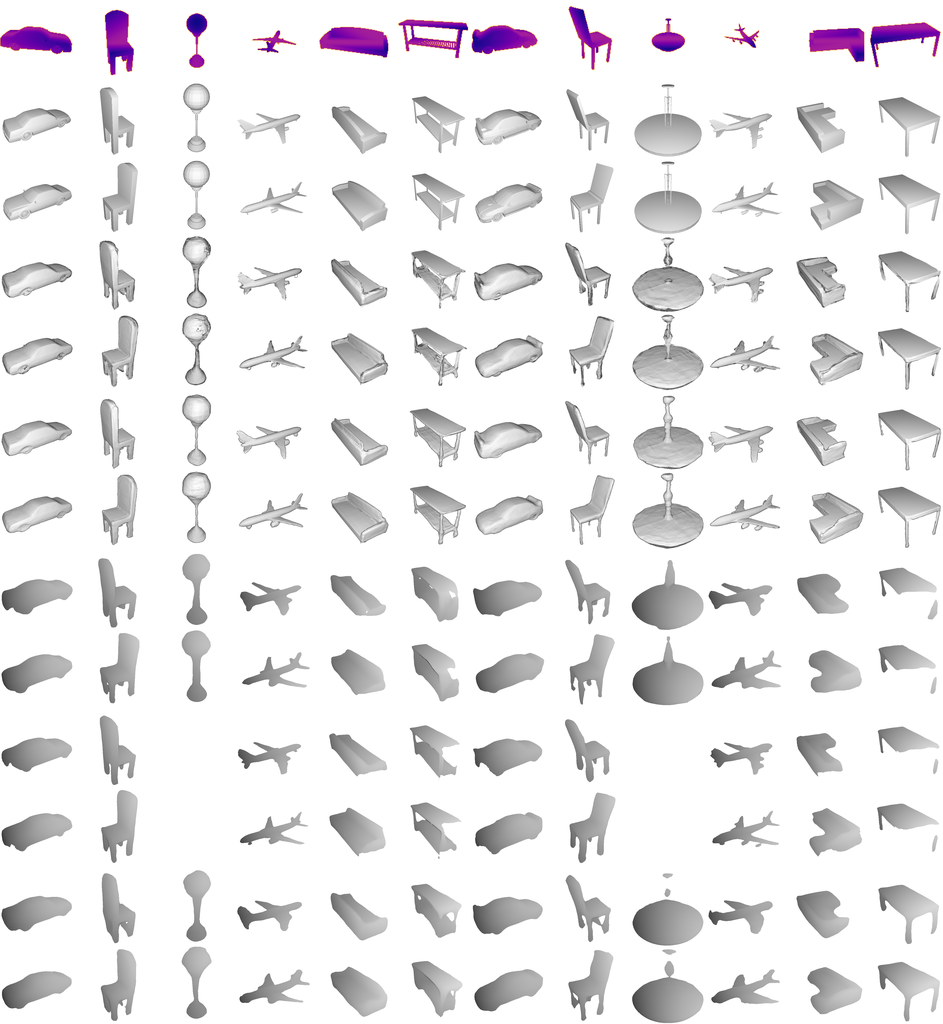}
        \put(-2,92){\rotatebox{90}{\small Given Depth}}
        
        \put(-3,80){\rotatebox{90}{\small Ground Truth}}
        \put(-1.2,78){\rotatebox{90}{\small View1}}
        \put(-1.2,86){\rotatebox{90}{\small View2}}

        \put(0,77.05){\linethickness{0.2mm}\color{black}\line(1,0){93}}
        \put(-3,62){\rotatebox{90}{\small DDF Renders (Ours)}}
        \put(-1.2,64){\rotatebox{90}{\small View1}}
        \put(-1.2,71){\rotatebox{90}{\small View2}}

        \put(-3,51){\rotatebox{90}{\small Ours-SDF}}
        \put(-1.2,48){\rotatebox{90}{\small View1}}
        \put(-1.2,55){\rotatebox{90}{\small View2}}

        \put(0,46.1){\linethickness{0.2mm}\color{black}\line(1,0){93}}
        \put(-3,34){\rotatebox{90}{\small PRIF Renders}}
        \put(-1.2,32){\rotatebox{90}{\small View1}}
        \put(-1.2,40){\rotatebox{90}{\small View2}}
        
        \put(0,30.8){\linethickness{0.2mm}\color{black}\line(1,0){93}}
        \put(-3,15){\rotatebox{90}{\small Depth-LFN Renders}}
        \put(-1.2,17){\rotatebox{90}{\small View1}}
        \put(-1.2,24){\rotatebox{90}{\small View2}}
        \put(0,15.5){\linethickness{0.2mm}\color{black}\line(1,0){93}}
        \put(-3,0){\rotatebox{90}{\small NeuralODF Renders}}
        \put(-1.2,1){\rotatebox{90}{\small View1}}
        \put(-1.2,9){\rotatebox{90}{\small View2}}

    \end{overpic}
    \caption{Qualitative results of model evaluation against PRIF, Depth-LFN, and NeuralODF on shape reconstruction from a given depth map. Each column shows reconstruction results for different shapes. Top rows: given depth map (top), ground truth geometry for reference (bottom). Middle-top rows: views rendered with $1$ forward pass from our DDF model (top) and views of 3D shapes reconstructed from our SDF model (bottom). Middle rows: views rendered with $1$ forward pass from PRIF model. Middle-bottom rows: views rendered with $1$ forward pass from Depth-LFN model. Bottom rows: views rendered with $1$ forward pass from NeuralODF model. We shade competitive models by estimating normals from rendered depth. As can be seen, our model results in more view-consistent shapes with better geometric details, given that we couple our DDF model with the SDF model. (Depth-LFN did not converge for lamps class.)}
    \label{fig:prifcomparison}
    \vspace{-0.5cm}
\end{figure*}

\textbf{Reconstruction from real-world videos:} We show qualitative results on reconstruction from $40$ real-world videos of chairs from the Redwood dataset~\cite{choi2015redwood} in Fig.~\ref{fig:mvQualResReal}. We optimize for the photometric consistency across frames using the depth predicted by our model. As the dataset consists of reconstructed meshes of entire scenes, we can evaluate the distance from the predicted shape using our model to the ground mesh. The mean distance is $4.0$cm.

\subsection{Shape from Silhouettes} \label{sec:silRec}
We evaluate our method on the challenging task of reconstructing 3D shapes from single-view silhouettes with a given camera pose. The task is much more challenging compared with reconstructing from single-view depth maps as we have no information about the shape of an object apart from a silhouette in an image.

We optimize for shape with the 3D reconstruction from the single-view depth maps algorithm, presented in Sec. 3.4 of the main paper, without the depth loss, i.e., $w_D=0.0$. We compare with DIST~\cite{liu2020dist} for this task while setting the weight of the depth loss in DIST's reconstruction algorithm to $0.0$. We test on the same test split as in 3D reconstruction from depth maps, i.e., the first $200$ shapes from test splits of each category and the first image from the test dataset of 3D-R2N2~\cite{choy20163d}.

\begin{figure*}[t]
    \centering
    
    \begin{overpic}[width=\linewidth,,tics=5]
        {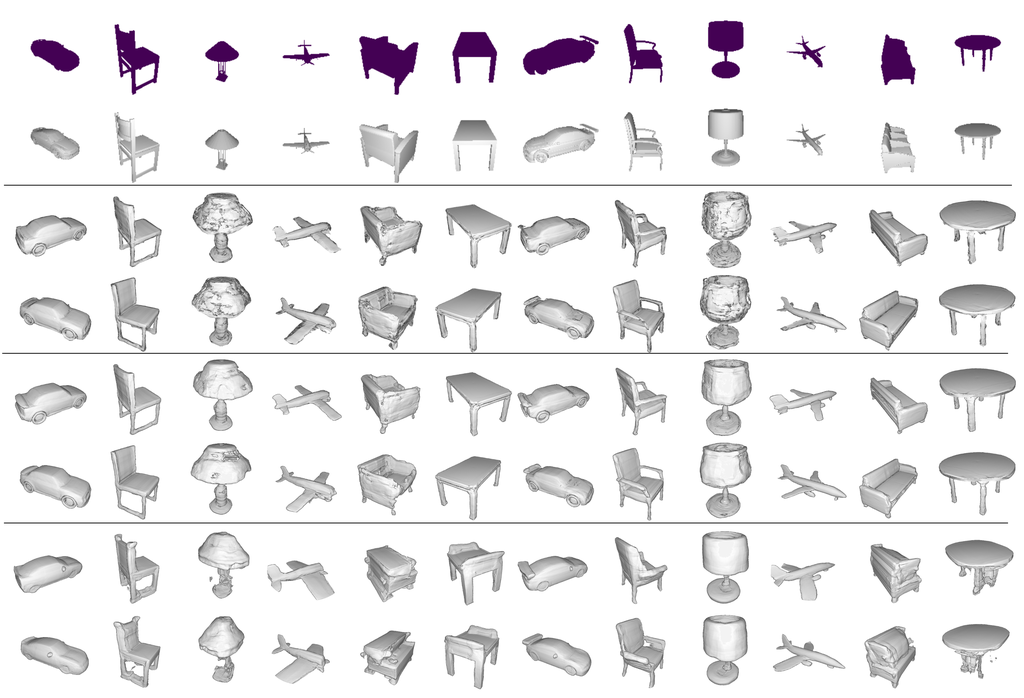}        
        \put(-3,53){\rotatebox{90}{\small Ground Truth}}
        \put(-1,52){\rotatebox{90}{\small Mesh}}
        \put(-1,59){\rotatebox{90}{\small Silhouette}}

        \put(-3,34){\rotatebox{90}{\small DDF Renders (Ours)}}
        \put(-1,44){\rotatebox{90}{\small View1}}
        \put(-1,35){\rotatebox{90}{\small View2}}

        \put(-3,23){\rotatebox{90}{\small Ours}}
        \put(-1,27.5){\rotatebox{90}{\small View1}}
        \put(-1,19){\rotatebox{90}{\small View2}}

        \put(-3,6){\rotatebox{90}{\small DIST}}
        \put(-1,10.5){\rotatebox{90}{\small View1}}
        \put(-1,2){\rotatebox{90}{\small View2}}
    \end{overpic}
    \caption{Qualitative results of 3D shapes reconstructed from a given silhouette. Each column shows reconstruction results for different shapes. Top rows: given silhouettes and meshes (for reference). Upper-middle rows: reconstructed views rendered with $1$ forward pass of our DDF model per camera ray. Middle rows: views of 3D shapes reconstructed by our SDF model. Last rows: views of 3D shape reconstructed using DIST~\cite{liu2020dist}.}
    \label{fig:comparisonsSil}
\end{figure*}

\begin{figure*}[t]
    \centering
    
    \begin{overpic}[width=\linewidth,,tics=5]
        {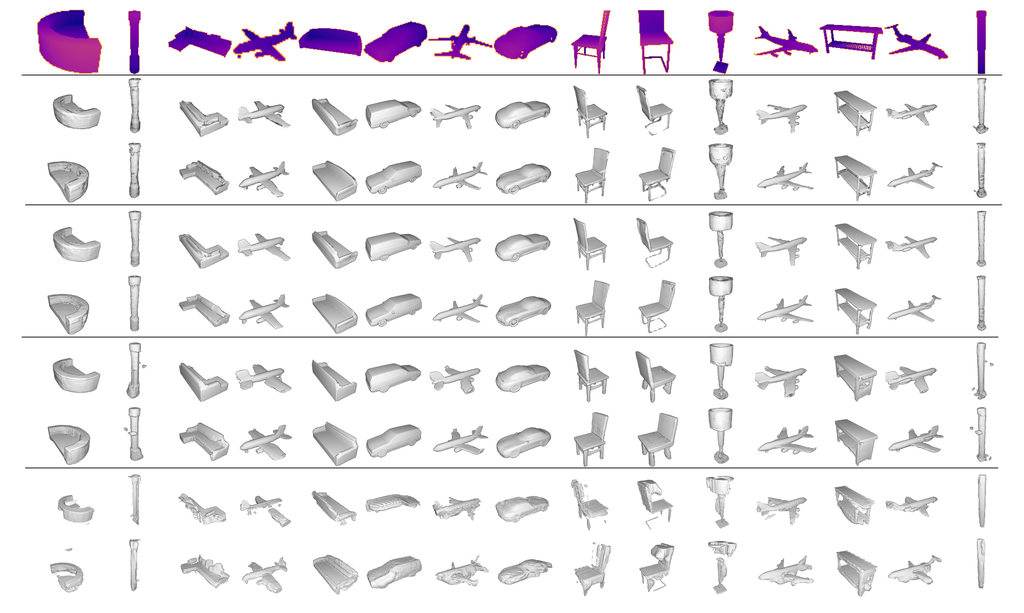}
        \put(-3,53.6){\rotatebox{90}{\small Ground}}
        \put(-1,54.1){\rotatebox{90}{\small Truth}}
        \put(1,54){\rotatebox{90}{\small Depth}}
        
        \put(-3,42){\rotatebox{90}{\small DDF Renders}}
        \put(-1,44){\rotatebox{90}{\small (Ours)}}
        \put(1,47){\rotatebox{90}{\small View1}}
        \put(1,41){\rotatebox{90}{\small View2}}

        \put(-1,31){\rotatebox{90}{\small Ours}}
        \put(1,34){\rotatebox{90}{\small View1}}
        \put(1,28){\rotatebox{90}{\small View2}}

        \put(-1,18){\rotatebox{90}{\small DIST}}
        \put(1,21){\rotatebox{90}{\small View1}}
        \put(1,15){\rotatebox{90}{\small View2}}

        \put(-1,5){\rotatebox{90}{\small IF-NET}}
        \put(1,8){\rotatebox{90}{\small View1}}
        \put(1,2){\rotatebox{90}{\small View2}}
    \end{overpic}
    \begin{overpic}[width=\linewidth,,tics=5]
        {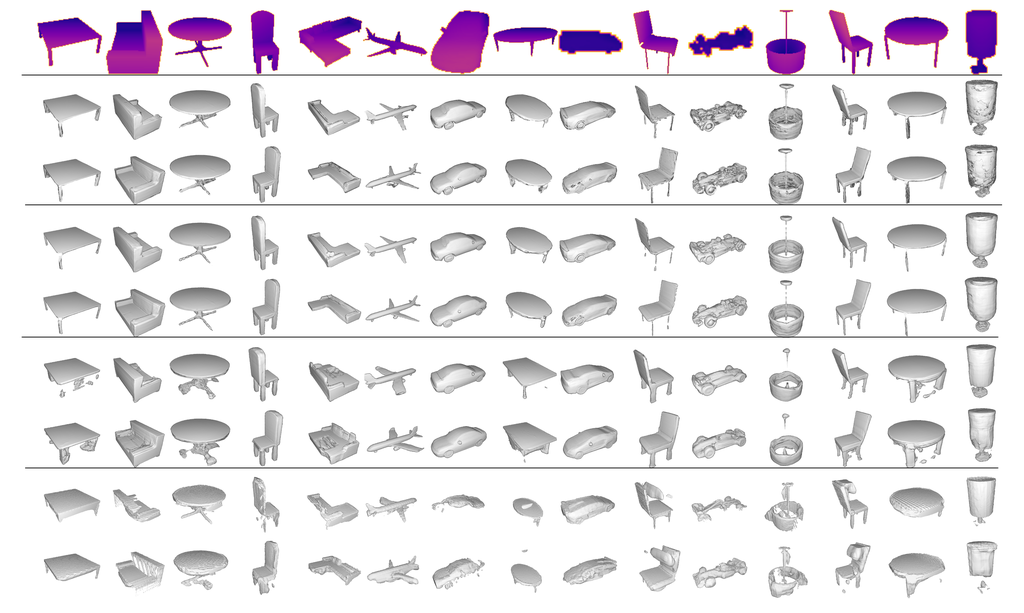}
        \put(-3,53.6){\rotatebox{90}{\small Ground}}
        \put(-1,54.1){\rotatebox{90}{\small Truth}}
        \put(1,54){\rotatebox{90}{\small Depth}}
        
        \put(-3,42){\rotatebox{90}{\small DDF Renders}}
        \put(-1,44){\rotatebox{90}{\small (Ours)}}
        \put(1,47){\rotatebox{90}{\small View1}}
        \put(1,41){\rotatebox{90}{\small View2}}

        \put(-1,31){\rotatebox{90}{\small Ours}}
        \put(1,34){\rotatebox{90}{\small View1}}
        \put(1,28){\rotatebox{90}{\small View2}}

        \put(-1,18){\rotatebox{90}{\small DIST}}
        \put(1,21){\rotatebox{90}{\small View1}}
        \put(1,15){\rotatebox{90}{\small View2}}

        \put(-1,5){\rotatebox{90}{\small IF-NET}}
        \put(1,8){\rotatebox{90}{\small View1}}
        \put(1,2){\rotatebox{90}{\small View2}}
    \end{overpic}
    \caption{Additional qualitative results of 3D reconstruction from single-view depth maps discussed in Sec. 4.2 of the main paper. Each column shows reconstruction results for different shapes. Top row: given depth map. Upper-middle rows: views rendered with $1$ forward pass from our DDF model. Middle rows: views of 3D shapes reconstructed by our SDF model. Lower-middle rows: views of 3D shape reconstructed using DIST~\cite{liu2020dist}. Last rows: views of 3D shapes reconstructed using IF-NET~\cite{chibane2020ifnet}.}
    \label{fig:comparisonSupp}
\end{figure*}

\textbf{Results}: We show qualitative results of reconstructing 3D shape from a silhouette in Fig.~\ref{fig:comparisonsSil}. We show the quantitative results in Tab.~\ref{tab:comparisonsSilTab}. We report $1000\times$ the L2 chamfer distance (see Sec. 4.3, main paper) between the reconstructed and ground truth meshes. Our method outperforms DIST~\cite{liu2020dist} in all classes. As can be seen, our algorithm reconstructs more plausible shapes as our DDF model's ray hit output can be used to fit our models to the given silhouettes, as corroborated by the reconstruction accuracy of DDF. Further, as our DDF model is trained using the Track-SDF loss and as our models share a latent space, our SDF reconstructions are more accurate compared with DIST's reconstructions.

\begin{table}[t]
    \centering
    \begin{tabular}{l|c|c|c||c|c|c}
                         &  \textbf{Ours} & \textbf{Ours} &  \textbf{DIST}  &  \textbf{Ours} & \textbf{Ours} &  \textbf{DIST} \\
                         &                & \textbf{DDF}  &                 &                & \textbf{DDF}  &                \\ \hline
                         & \multicolumn{3}{c||}{$1000\times$ CD $\downarrow$ (Mean)}& \multicolumn{3}{c}{$1000\times$ CD $\downarrow$ (Median)}\\ \hline
         Car   & 0.99          & \textbf{0.75}  & 1.53  & 0.90          & \textbf{0.51}  & 1.16 \\ 
         Chair & 2.70          & \textbf{2.06}  & 7.84  & 1.93          & \textbf{1.51}  & 4.58 \\ 
         Lamp  & \textbf{6.91} & 8.02           & 11.92 & \textbf{3.81} & 4.33           & 5.19 \\ 
         Plane & \textbf{0.74} & 0.78           & 5.74  & 0.45          & \textbf{0.40}  & 4.22 \\ 
         Sofa  & \textbf{2.08} & 2.33           & 3.82  & 1.69          & \textbf{1.62}  & 2.18 \\  
         Table & 3.25          & \textbf{1.97}  & 5.95  & 2.37          & \textbf{1.28}  & 3.35 \\ 
    \end{tabular}
    \caption{Quantitative results on 3D shape reconstruction given a silhouette. Our method outperforms DIST~\cite{liu2020dist} in all the classes by a large margin as our DDF model predicts silhouette along with directional distance. Further, as the DDF model is trained to track the surface predicted by SDF model, both our DDF and SDF models can reconstruct from silhouettes better than the state-of-the-art.
    }
    \label{tab:comparisonsSilTab}
\end{table}
\section{Visualizations}\label{sec:visuals}

For the results named ``DDF Renders'' or ``DDF Reconstructions'', we visualize the results of our method with one evaluation of DDF per ray. We obtain the silhouette predicted by the DDF model and use the SDF model to compute the surface normals for shading. The rendering time we reported in the main paper includes the time to compute normals using finite differences. We use sphere tracing to show our 3D reconstructions, and the results of DIST~\cite{liu2020dist}. We show reconstructed meshes for IF-NET~\cite{chibane2020ifnet}. We render meshes of IF-NETs for visualizations, and 3D-R2N2's test data set for obtaining depth maps and masks with the help of Trimesh~\cite{haggerty2019trimesh}. For shading, we use an accelerated version of Blinn-Phong shader from i3DMM's code~\cite{yenamandra2021i3dmm}.

{\small
\bibliographystyle{ieee_fullname}
\bibliography{egbib}

\begin{thebibliography}{10}\itemsep=-1pt

\bibitem{aumentado2021pddf}
Tristan Aumentado-Armstrong, Stavros Tsogkas, Sven Dickinson, and Allan~D. Jepson.
\newblock Representing 3d shapes with probabilistic directed distance fields.
\newblock In {\em Proceedings of the IEEE/CVF Conference on Computer Vision and Pattern Recognition (CVPR)}, pages 19343--19354, June 2022.

\bibitem{blanz1999bfm}
Volker Blanz and Thomas Vetter.
\newblock A morphable model for the synthesis of 3d faces.
\newblock In {\em Proceedings of the 26th Annual Conference on Computer Graphics and Interactive Techniques}, 1999.

\bibitem{bogo2016keepitsmpl}
Federica Bogo, Angjoo Kanazawa, Christoph Lassner, Peter Gehler, Javier Romero, and Michael~J. Black.
\newblock Keep it {SMPL}: Automatic estimation of {3D} human pose and shape from a single image.
\newblock In {\em European Conference on Computer Vision}, 2016.

\bibitem{chabra2020deep}
Rohan Chabra, Jan~E Lenssen, Eddy Ilg, Tanner Schmidt, Julian Straub, Steven Lovegrove, and Richard Newcombe.
\newblock Deep local shapes: Learning local sdf priors for detailed 3d reconstruction.
\newblock In {\em European Conference on Computer Vision}, pages 608--625. Springer, 2020.

\bibitem{chanmonteiro2020pigan}
Eric Chan, Marco Monteiro, Petr Kellnhofer, Jiajun Wu, and Gordon Wetzstein.
\newblock pi-gan: Periodic implicit generative adversarial networks for 3d-aware image synthesis.
\newblock In {\em arXiv}, 2020.

\bibitem{chan2022eg3d}
Eric~R. Chan, Connor~Z. Lin, Matthew~A. Chan, Koki Nagano, Boxiao Pan, Shalini~De Mello, Orazio Gallo, Leonidas Guibas, Jonathan Tremblay, Sameh Khamis, Tero Karras, and Gordon Wetzstein.
\newblock Efficient geometry-aware {3D} generative adversarial networks.
\newblock In {\em CVPR}, 2022.

\bibitem{chang2015shapenet}
Angel~X. Chang, Thomas Funkhouser, Leonidas Guibas, Pat Hanrahan, Qixing Huang, Zimo Li, Silvio Savarese, Manolis Savva, Shuran Song, Hao Su, Jianxiong Xiao, Li Yi, and Fisher Yu.
\newblock Shapenet: An information-rich 3d model repository, 2015.

\bibitem{chen2022tensorf}
Anpei Chen, Zexiang Xu, Andreas Geiger, Jingyi Yu, and Hao Su.
\newblock Tensorf: Tensorial radiance fields.
\newblock In {\em European Conference on Computer Vision (ECCV)}, 2022.

\bibitem{chen2018implicitdecoder}
Zhiqin Chen and Hao Zhang.
\newblock Learning implicit fields for generative shape modeling.
\newblock {\em Proceedings of IEEE Conference on Computer Vision and Pattern Recognition (CVPR)}, 2019.

\bibitem{chibane2020ifnet}
Julian Chibane, Thiemo Alldieck, and Gerard Pons-Moll.
\newblock Implicit functions in feature space for 3d shape reconstruction and completion.
\newblock In {\em {IEEE} Conference on Computer Vision and Pattern Recognition (CVPR)}. {IEEE}, jun 2020.

\bibitem{choi2015redwood}
Sungjoon Choi, Qian-Yi Zhou, and Vladlen Koltun.
\newblock Robust reconstruction of indoor scenes.
\newblock In {\em IEEE Conference on Computer Vision and Pattern Recognition (CVPR)}, 2015.

\bibitem{choy20163d}
Christopher~B Choy, Danfei Xu, JunYoung Gwak, Kevin Chen, and Silvio Savarese.
\newblock 3d-r2n2: A unified approach for single and multi-view 3d object reconstruction.
\newblock In {\em Proceedings of the European Conference on Computer Vision ({ECCV})}, 2016.

\bibitem{cremers2006dynamical}
Daniel Cremers.
\newblock Dynamical statistical shape priors for level set-based tracking.
\newblock {\em IEEE Transactions on Pattern Analysis and Machine Intelligence}, 28(8):1262--1273, 2006.

\bibitem{haggerty2019trimesh}
{Dawson-Haggerty et al.}
\newblock trimesh.

\bibitem{fang2022tineuvox}
Jiemin Fang, Taoran Yi, Xinggang Wang, Lingxi Xie, Xiaopeng Zhang, Wenyu Liu, Matthias Nie{\ss}ner, and Qi Tian.
\newblock Fast dynamic radiance fields with time-aware neural voxels.
\newblock {\em arxiv:2205.15285}, 2022.

\bibitem{Faugeras-Keriven-98}
O. Faugeras and R. Keriven.
\newblock Variational principles, surface evolution, {PDE}'s, level set methods, and the stereo problem.
\newblock {\em IEEE TIP}, 7(3):336--344, Mar. 1998.

\bibitem{feng2022prif}
Brandon~Y. Feng, Yinda Zhang, Danhang Tang, Ruofei Du, and Amitabh Varshney.
\newblock Prif: Primary ray-based implicit function.
\newblock In {\em Proceedings of the European Conference on Computer Vision (ECCV)}, October 2022.

\bibitem{gkioxari2019meshrcnn}
Justin~Johnson Georgia~Gkioxari, Jitendra~Malik.
\newblock Mesh r-cnn.
\newblock {\em Proceedings of the IEEE International Conference on Computer Vision}, 2019.

\bibitem{hart1996spheretracing}
John~C. Hart.
\newblock Sphere tracing: a geometric method for the antialiased ray tracing of implicit surfaces.
\newblock {\em The Visual Computer}, 12(10):527--545, Dec. 1996.

\bibitem{houchens2022neuralodf}
Trevor Houchens, Cheng-You Lu, Shivam Duggal, Rao Fu, and Srinath Sridhar.
\newblock Neuralodf: Learning omnidirectional distance fields for 3d shape representation, 2022.

\bibitem{jiang2020local}
Chiyu Jiang, Avneesh Sud, Ameesh Makadia, Jingwei Huang, Matthias Nie{\ss}ner, Thomas Funkhouser, et~al.
\newblock Local implicit grid representations for 3d scenes.
\newblock In {\em Proceedings of the IEEE/CVF Conference on Computer Vision and Pattern Recognition}, pages 6001--6010, 2020.

\bibitem{kingma2014adam}
Diederik~P. Kingma and Jimmy Ba.
\newblock Adam: A method for stochastic optimization, 2014.

\bibitem{kobbelt2001featureSensitive}
Leif~P. Kobbelt, Mario Botsch, Ulrich Schwanecke, and Hans-Peter Seidel.
\newblock Feature sensitive surface extraction from volume data.
\newblock In {\em Proceedings of the 28th Annual Conference on Computer Graphics and Interactive Techniques}, SIGGRAPH '01, page 57–66, New York, NY, USA, 2001. Association for Computing Machinery.

\bibitem{kohlberger20064d}
Timo Kohlberger, Daniel Cremers, Mika{\"e}l Rousson, Ramamani Ramaraj, and Gareth Funka-Lea.
\newblock 4d shape priors for a level set segmentation of the left myocardium in spect sequences.
\newblock In {\em International Conference on Medical Image Computing and Computer-Assisted Intervention}, pages 92--100. Springer, 2006.

\bibitem{leventon2002statistical}
Michael~E Leventon, W~Eric~L Grimson, and Olivier Faugeras.
\newblock Statistical shape influence in geodesic active contours.
\newblock In {\em 5th IEEE EMBS International Summer School on Biomedical Imaging, 2002.}, pages 8--pp. IEEE, 2002.

\bibitem{lin2019photometric}
Chen-Hsuan Lin, Oliver Wang, Bryan~C Russell, Eli Shechtman, Vladimir~G Kim, Matthew Fisher, and Simon Lucey.
\newblock Photometric mesh optimization for video-aligned 3d object reconstruction.
\newblock In {\em IEEE Conference on Computer Vision and Pattern Recognition ({CVPR})}, 2019.

\bibitem{liu2020dist}
Shaohui Liu, Yinda Zhang, Songyou Peng, Boxin Shi, Marc Pollefeys, and Zhaopeng Cui.
\newblock Dist: Rendering deep implicit signed distance function with differentiable sphere tracing.
\newblock In {\em IEEE Conference on Computer Vision and Pattern Recognition (CVPR)}, 2020.

\bibitem{loper2015smpl}
Matthew Loper, Naureen Mahmood, Javier Romero, Gerard Pons-Moll, and Michael~J. Black.
\newblock {SMPL}: A skinned multi-person linear model.
\newblock {\em ACM Transactions on Graphics}, Oct. 2015.

\bibitem{mescheder2019onet}
Lars Mescheder, Michael Oechsle, Michael Niemeyer, Sebastian Nowozin, and Andreas Geiger.
\newblock Occupancy networks: Learning 3d reconstruction in function space.
\newblock In {\em Proceedings IEEE Conf. on Computer Vision and Pattern Recognition (CVPR)}, 2019.

\bibitem{mildenhall2020nerf}
Ben Mildenhall, Pratul~P. Srinivasan, Matthew Tancik, Jonathan~T. Barron, Ravi Ramamoorthi, and Ren Ng.
\newblock Nerf: Representing scenes as neural radiance fields for view synthesis.
\newblock In {\em ECCV}, 2020.

\bibitem{mueller2022instant}
Thomas M\"uller, Alex Evans, Christoph Schied, and Alexander Keller.
\newblock Instant neural graphics primitives with a multiresolution hash encoding.
\newblock {\em ACM Trans. Graph.}, 41(4):102:1--102:15, July 2022.

\bibitem{muraki1991volumetric}
Shigeru Muraki.
\newblock Volumetric shape description of range data using “blobby model”.
\newblock In {\em Proceedings of the 18th annual conference on Computer graphics and interactive techniques}, pages 227--235, 1991.

\bibitem{neff2021donerf}
Thomas Neff, Pascal Stadlbauer, Mathias Parger, Andreas Kurz, Joerg~H. Mueller, Chakravarty R.~Alla Chaitanya, Anton~S. Kaplanyan, and Markus Steinberger.
\newblock {DONeRF: Towards Real-Time Rendering of Compact Neural Radiance Fields using Depth Oracle Networks}.
\newblock {\em Computer Graphics Forum}, 40(4), 2021.

\bibitem{niemeyer2020dvr}
Michael Niemeyer, Lars Mescheder, Michael Oechsle, and Andreas Geiger.
\newblock Differentiable volumetric rendering: Learning implicit 3d representations without 3d supervision.
\newblock In {\em Proc. IEEE Conf. on Computer Vision and Pattern Recognition (CVPR)}, 2020.

\bibitem{park2019deepsdf}
Jeong~Joon Park, Peter Florence, Julian Straub, Richard Newcombe, and Steven Lovegrove.
\newblock Deepsdf: Learning continuous signed distance functions for shape representation.
\newblock In {\em The IEEE Conference on Computer Vision and Pattern Recognition (CVPR)}, June 2019.

\bibitem{paszke2019pytorch}
Adam Paszke, Sam Gross, Francisco Massa, Adam Lerer, James Bradbury, Gregory Chanan, Trevor Killeen, Zeming Lin, Natalia Gimelshein, Luca Antiga, Alban Desmaison, Andreas Köpf, Edward Yang, Zach DeVito, Martin Raison, Alykhan Tejani, Sasank Chilamkurthy, Benoit Steiner, Lu Fang, Junjie Bai, and Soumith Chintala.
\newblock Pytorch: An imperative style, high-performance deep learning library, 2019.

\bibitem{peng2020convolutional}
Songyou Peng, Michael Niemeyer, Lars Mescheder, Marc Pollefeys, and Andreas Geiger.
\newblock Convolutional occupancy networks.
\newblock In {\em European Conference on Computer Vision}, pages 523--540. Springer, 2020.

\bibitem{piala2021terminerf}
Martin Piala and Ronald Clark.
\newblock Terminerf: Ray termination prediction for efficient neural rendering.
\newblock In {\em 2021 International Conference on 3D Vision (3DV)}, pages 1106--1114. IEEE, 2021.

\bibitem{ricci1973constructive}
Antonio Ricci.
\newblock A constructive geometry for computer graphics.
\newblock {\em The Computer Journal}, 16(2):157--160, 1973.

\bibitem{saito20183d}
Shunsuke Saito, Liwen Hu, Chongyang Ma, Hikaru Ibayashi, Linjie Luo, and Hao Li.
\newblock 3d hair synthesis using volumetric variational autoencoders.
\newblock {\em ACM Transactions on Graphics (TOG)}, 37(6):1--12, 2018.

\bibitem{saito2019pifu}
Shunsuke Saito, Zeng Huang, Ryota Natsume, Shigeo Morishima, Angjoo Kanazawa, and Hao Li.
\newblock Pifu: Pixel-aligned implicit function for high-resolution clothed human digitization.
\newblock In {\em Proceedings of the IEEE/CVF International Conference on Computer Vision}, pages 2304--2314, 2019.

\bibitem{saito2020pifuhd}
Shunsuke Saito, Tomas Simon, Jason Saragih, and Hanbyul Joo.
\newblock Pifuhd: Multi-level pixel-aligned implicit function for high-resolution 3d human digitization.
\newblock In {\em Proceedings of the IEEE/CVF Conference on Computer Vision and Pattern Recognition}, pages 84--93, 2020.

\bibitem{yu_and_fridovichkeil2021plenoxels}
{Sara Fridovich-Keil and Alex Yu}, Matthew Tancik, Qinhong Chen, Benjamin Recht, and Angjoo Kanazawa.
\newblock Plenoxels: Radiance fields without neural networks.
\newblock In {\em CVPR}, 2022.

\bibitem{sitzmann2021lfns}
Vincent Sitzmann, Semon Rezchikov, William~T. Freeman, Joshua~B. Tenenbaum, and Fredo Durand.
\newblock Light field networks: Neural scene representations with single-evaluation rendering.
\newblock In {\em Proc. NeurIPS}, 2021.

\bibitem{sitzmann2019scene}
Vincent Sitzmann, Michael Zollh{\"o}fer, and Gordon Wetzstein.
\newblock Scene representation networks: Continuous 3d-structure-aware neural scene representations.
\newblock {\em Advances in Neural Information Processing Systems}, 32, 2019.

\bibitem{sommer2022gradientsdf}
C Sommer, L Sang, D Schubert, and D Cremers.
\newblock Gradient-sdf: {A} semi-implicit surface representation for 3d reconstruction.
\newblock {\em arXiv preprint}, 2021.

\bibitem{sturm2013copyme3d}
J{\"u}rgen Sturm, Erik Bylow, Fredrik Kahl, and Daniel Cremers.
\newblock Copyme3d: Scanning and printing persons in 3d.
\newblock In {\em German Conference on Pattern Recognition}, pages 405--414. Springer, 2013.

\bibitem{takikawa2021nglod}
Towaki Takikawa, Joey Litalien, Kangxue Yin, Karsten Kreis, Charles Loop, Derek Nowrouzezahrai, Alec Jacobson, Morgan McGuire, and Sanja Fidler.
\newblock Neural geometric level of detail: Real-time rendering with implicit 3d shapes.
\newblock In {\em Proceedings of the IEEE/CVF Conference on Computer Vision and Pattern Recognition (CVPR)}, pages 11358--11367, June 2021.

\bibitem{tewari17MoFA}
Ayush Tewari, Michael Zoll{\"o}fer, Hyeongwoo Kim, Pablo Garrido, Florian Bernard, Patrick Perez, and Theobalt Christian.
\newblock {MoFA: Model-based Deep Convolutional Face Autoencoder for Unsupervised Monocular Reconstruction}.
\newblock In {\em Proceedings of the IEEE International Conference on Computer Vision}, 2017.

\bibitem{tiwari22posendf}
Garvita Tiwari, Dimitrije Antic, Jan~Eric Lenssen, Nikolaos Sarafianos, Tony Tung, and Gerard Pons-Moll.
\newblock Pose-ndf: Modeling human pose manifolds with neural distance fields.
\newblock In {\em European Conference on Computer Vision ({ECCV})}, October 2022.

\bibitem{tiwari21neuralgif}
Garvita Tiwari, Nikolaos Sarafianos, Tony Tung, and Gerard Pons-Moll.
\newblock Neural-gif: Neural generalized implicit functions for animating people in clothing.
\newblock In {\em International Conference on Computer Vision ({ICCV})}, October 2021.

\bibitem{toeppe_et_al_cvpr13}
Eno Toeppe, Claudia Nieuwenhuis, and Daniel Cremers.
\newblock Volume constraints for single view reconstruction.
\newblock In {\em Proceedings of the IEEE Conference on Computer Vision and Pattern Recognition}, 2013.

\bibitem{wang2020directshape}
Rui Wang, Nan Yang, Joerg Stueckler, and Daniel Cremers.
\newblock Directshape: Photometric alignment of shape priors for visual vehicle pose and shape estimation.
\newblock In {\em Proceedings of the IEEE International Conference on Robotics and Automation}, 2020.

\bibitem{Wu_2020_CVPR}
Shangzhe Wu, Christian Rupprecht, and Andrea Vedaldi.
\newblock Unsupervised learning of probably symmetric deformable 3d objects from images in the wild.
\newblock In {\em Proceedings of the IEEE Conference on Computer Vision and Pattern Recognition}, 2020.

\bibitem{xu2019disn}
Qiangeng Xu, Weiyue Wang, Duygu Ceylan, Radomir Mech, and Ulrich Neumann.
\newblock Disn: Deep implicit surface network for high-quality single-view 3d reconstruction.
\newblock In H. Wallach, H. Larochelle, A. Beygelzimer, F. d\textquotesingle Alch\'{e}-Buc, E. Fox, and R. Garnett, editors, {\em Advances in Neural Information Processing Systems 32}, pages 492--502. Curran Associates, Inc., 2019.

\bibitem{yariv2020idr}
Lior Yariv, Yoni Kasten, Dror Moran, Meirav Galun, Matan Atzmon, Basri Ronen, and Yaron Lipman.
\newblock Multiview neural surface reconstruction by disentangling geometry and appearance.
\newblock {\em Advances in Neural Information Processing Systems}, 33, 2020.

\bibitem{ye2021joint}
Zhenzhang Ye, Tarun Yenamandra, Florian Bernard, and Daniel Cremers.
\newblock Joint deep multi-graph matching and 3d geometry learning from inhomogeneous 2d image collections.
\newblock In {\em AAAI}, 2022.

\bibitem{yenamandra2021i3dmm}
Tarun Yenamandra, Ayush Tewari, Florian Bernard, Hans-Peter Seidel, Mohamed Elgharib, Daniel Cremers, and Christian Theobalt.
\newblock i3dmm: Deep implicit 3d morphable model of human heads.
\newblock In {\em Proceedings of the IEEE/CVF Conference on Computer Vision and Pattern Recognition}, pages 12803--12813, 2021.

\bibitem{yu2020pixelnerf}
Alex Yu, Vickie Ye, Matthew Tancik, and Angjoo Kanazawa.
\newblock pixelnerf: Neural radiance fields from one or few images, 2020.

\bibitem{zobeidi2021sddf}
Ehsan Zobeidi and Nikolay Atanasov.
\newblock A deep signed directional distance function for object shape representation.
\newblock {\em arXiv preprint arXiv:2107.11024}, 2021.

\end{thebibliography}
}
\end{document}